%% file: NRP_Paper.tex
\documentclass[10pt,twocolumn,letterpaper]{article}

\usepackage{cvpr}
\usepackage{times}
\usepackage{epsfig}
\usepackage{graphicx}
\usepackage{amsmath}
\usepackage{amssymb}

\usepackage{booktabs}
\usepackage{bm}
\usepackage{algorithm, algpseudocode}

\usepackage{textcomp}
\usepackage{multirow}
\usepackage[outercaption]{sidecap}
\usepackage{makecell}
\usepackage{stmaryrd}
\usepackage{wrapfig}
\usepackage{caption}
\usepackage{enumitem}
\usepackage{subcaption}
\usepackage{pifont}  
\usepackage{array}
\usepackage{enumitem}
\usepackage{pifont}

\newcommand{\PreserveBackslash}[1]{\let\temp=\\#1\let\\=\temp}
\newcolumntype{C}[1]{>{\PreserveBackslash\centering}p{#1}}
\newcolumntype{R}[1]{>{\PreserveBackslash\raggedleft}p{#1}}
\newcolumntype{L}[1]{>{\PreserveBackslash\raggedright}p{#1}}

\newcommand{\cmark}{\ding{51}}%
\newcommand{\xmark}{\ding{55}}%

\usepackage[pagebackref=true,breaklinks=true,letterpaper=true,colorlinks,bookmarks=false]{hyperref}

\newcommand{\F}{{\mathcal{F}_{\psi}}}

\cvprfinalcopy % *** Uncomment this line for the final submission

\def\cvprPaperID{8087} % *** Enter the CVPR Paper ID here

\ifcvprfinal\fi
\begin{document}

%%%%%%%%% TITLE
\title{A Self-supervised Approach for Adversarial Robustness}
% Towards Adversarial Robustness with Self-supervision
\author{
    Muzammal Naseer$^{*\dagger\ddagger}$, Salman Khan$^{\dagger}$, Munawar Hayat$^\dagger$, 
    Fahad Shahbaz Khan$^{\dagger \mathsection}$, Fatih Porikli$^*$\\
  $^*$Australian National University, Australia, $^\ddagger$Data61-CSIRO, Australia\\
  $^\dagger$Inception Institute of Artificial Intelligence, UAE, $^\mathsection$CVL, Link\"oping University, Sweden\\
  \texttt{\small \{muzammal.naseer,fatih.porikli\}@anu.edu.au} \\
  \texttt{\small \{salman.khan,munawar.hayat,fahad.khan\}@inceptioniai.org} 
  }

\maketitle
%\thispagestyle{empty}

%-------------------------------------------------------------------------
\begin{abstract}
Adversarial examples can cause catastrophic mistakes in Deep Neural Network (DNNs)  based vision systems e.g., for classification, segmentation and object detection. The vulnerability of DNNs against such attacks can prove a major roadblock towards their real-world deployment. Transferability of adversarial examples demand generalizable defenses that can provide cross-task protection. Adversarial training that enhances robustness by modifying target model's parameters lacks such generalizability. On the other hand, different input processing based defenses fall short in the face of continuously evolving attacks. In this paper, we take the first step to combine the benefits of both approaches and propose a self-supervised adversarial training mechanism in the input space. By design, our defense is a generalizable approach and provides significant robustness against the \textbf{unseen} adversarial attacks (\eg by reducing the success rate of translation-invariant \textbf{ensemble} attack from 82.6\% to 31.9\% in comparison to previous state-of-the-art). It can be deployed as a plug-and-play solution to protect a variety of vision systems, as we demonstrate for the case of classification, segmentation and detection. Code is available at:  {\small\url{https://github.com/Muzammal-Naseer/NRP}}.
\end{abstract}

%-------------------------------------------------------------------------
\section{Introduction}
Adversarial training (AT) has shown great potential to safeguard neural networks from adversarial attacks \cite{madry2018towards, shafahi2019adversarial}. So far in literature, AT is performed in the model space i.e., a model's parameters are modified by minimizing empirical risk for a given data distribution as well as the perturbed images. Such AT strategy results in the following challenges. \textbf{\emph{(a) Task dependency:}} AT is task-dependent \eg robust classification models cannot directly be incorporated into an object detection or a segmentation pipeline, since the overall system would still require further training with modified task-dependant loss functions. \textbf{\emph{(b) Computational cost:}}  AT is computationally expensive \cite{madry2018towards} which restricts its applicability to high-dimensional and large-scale datasets such as ImageNet \cite{ILSVRC15}. \textbf{\emph{(c) Accuracy drop:}} models trained with AT lose significant accuracy on the original distribution \eg  ResNet50 \cite{he2016deep} accuracy on ImageNet validation set drops from 76\% to 64\% when robustified against PGD attack \cite{madry2018towards} at a perturbation budget of only $\epsilon \le 2$ (\ie maximum change in each pixel can be 2/255). \textbf{\emph{(d) Label leakage:}} supervised AT suffers from label leakage \cite{kurakin2016adversarial} which allows the model to overfit on perturbations thus affecting model generalization to unseen adversaries \cite{zhang2019defense}.

\begin{figure}[t]
\centering
  \begin{minipage}{.48\textwidth}
  	\centering
  	 % lelf lower right up 
    \includegraphics[width=\linewidth, keepaspectratio]{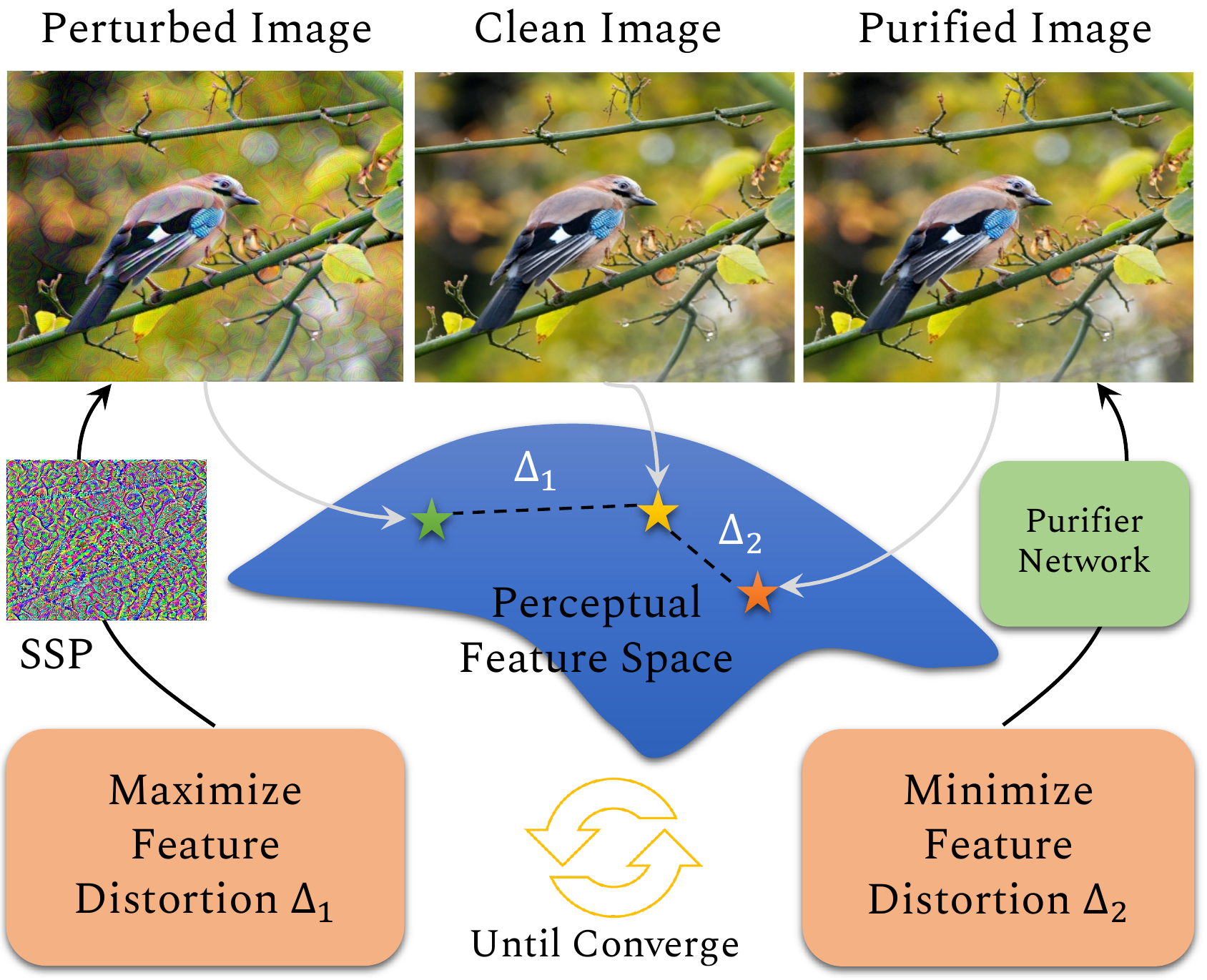}\
  \end{minipage}
  \caption{\small Our main idea is to train a Purifier Network in a self-supervised manner. We generate perturbed images using our proposed \emph{Self-supervised Perturbation} ({SSP}) attack that disrupts the deep perceptual features. The \emph{Purifier Network} projects back the perturbed images close to the perceptual space of clean images. This creates a training loop independent of the task or label space.}
\label{fig:main-concept-fig}
\end{figure}

In comparison to AT, input processing methods \cite{guo2017countering, xie2017mitigating} for adversarial defense are scalable and can work across different tasks. However, they have been broken in white-box settings \cite{athalye2018obfuscated} and shown to be least effective in black-box settings. For example, \cite{dong2019evading} successfully transfer their attack against multiple input processing based defenses even when the backbone architecture is adversarially trained using \cite{tramer2017ensemble}. Furthermore, input transformations (\eg, Gaussian smoothing and JPEG compression) can maximize the attack strength instead of minimizing it \cite{naseer2019cross, dong2019evading}.

Motivated by the complementary strengths of AT and input processing methods, we propose a self-supervised AT mechanism in the input space. Our approach (Fig.~\ref{fig:main-concept-fig}) uses a min-max (saddle point) formulation to learn an optimal  input processing function that enhances model robustness. In this way, our optimization rule implicitly performs AT. The main advantage of our approach is its generalization ability, once trained on a dataset, it can be applied off-the-shelf to safeguard a completely different model. This makes it a more attractive solution compared to popular AT approaches that are computationally expensive (and thus less scalable to large-scale datasets). Furthermore, in comparison to previous pre-processing based defenses that are found to be vulnerable towards recent attacks, our defense demonstrates better robustness.  Our main contributions are:\vspace{-0.5em}
\begin{itemize}[leftmargin=*]
\setlength{\itemsep}{-0.5em}
    \item \textbf{\emph{Task Generalizability:}} 
    To ensure a task independent AT mechanism, we propose to adversarially train a purifying model named Neural Representation Purifier (NRP). Once trained, NRP can be deployed to safeguard %any computer vision system 
    across different tasks, e.g., classification, detection and segmentation, without any additional training (Sec.~\ref{sec:NRP}).
    \item \textbf{\emph{Self-Supervision:}} The supervisory signal used for AT should be self-supervised to make it independent of label space. To this end,  we propose an algorithm to train NRP on adversaries found in the feature space in random directions to avoid any label leakage (Sec.~\ref{subsec:self-supervision}).
    \item \textbf{\emph{Defense against strong perturbations:}} Attacks are continuously evolving. In order for NRP to generalize, it should be trained on worst-case perturbations that are transferable across different tasks. We propose to find highly transferable perceptual adversaries  (Sec.~\ref{subsec:self-supervised-attack}). 
    %Significance of this choice is thoroughly discussed in Sec.\ref{subsec:self-supervision} and supplementary material.
    \item \textbf{\emph{Maintaining Accuracy:}} A strong defense must concurrently maintain accuracy on the original data distribution. We propose to train the NRP with an additional discriminator to bring adversarial examples close to original samples by recovering the fine texture details (Sec.~\ref{sec:def_res}).
\end{itemize}

%-------------------------------------------------------------------------
\section{Related Work}
\textbf{Defenses:}  A major class of adversarial defenses processes the input images to achieve robustness against adversarial patterns. For example, \cite{guo2017countering} used \textbf{JPEG} compression to remove high-frequency components that are less important to human vision using discrete cosine transform. A compressed sensing approach called Total Variation Minimization (\textbf{TVM}) was proposed in \cite{guo2017countering} to remove the small localized changes caused by adversarial perturbations. Xie \etal \cite{xie2018mitigating} introduced the process of Random Resizing and Padding (\textbf{R\&P}) as a pre-processing step to mitigate the adversarial effect. A High-level representation Guided Denoiser (\textbf{HGD}) \cite{Liao2017DefenseAA} framework was used as a pre-processing step to remove perturbations. NeurIPS 2017 Defense Competition Rank-3 (\textbf{NeurIPS-r3}) approach \cite{NIPS-r3} introduced a two step prep-processing pipeline where the images first undergo a series of transformations (JPEG, rotation, zoom, shift and sheer) and then passed through an ensemble of adversarially trained models to obtain the weighted output response as a prediction. \cite{shen2017ape} proposed to recover adversaries using GAN and \cite{mustafa2019image} super-resolve images to minimize adversarial effect.
As compared to the above defenses, we design an input processing model that derives a self-supervised signal from the deep feature space to adversarially train the defense model. Our results show significantly superior performance to all so-far developed input processing based defenses.

\textbf{Attacks:} The self-supervised perturbation signal obtained to adversarially train our proposed approach can also be used as an adversarial attack. Since the seminal work of Szegedy \etal \cite{szegedy2013intriguing}, many adversarial attack algorithms \cite{goodfellow2014explaining, Kurakin2016AdversarialEI, athalye2017synthesizing, Dong_2018_CVPR} have been proposed to show the vulnerability of neural networks against imperceptible changes to inputs. A single-step attack, called Fast Gradient Sign Method (\textbf{FGSM}), was proposed in \cite{goodfellow2014explaining}. In a follow-up work, Kurakin \etal \cite{Kurakin2016AdversarialEI} proposed a robust multi-step attack, called Iterative Fast Gradient Sign Method (\textbf{I-FGSM}) that iteratively searches the loss surface of a network under a given metric norm. 
To improve transferability, a variant of I-FGSM, called momentum iterative fast gradient sign method (\textbf{MI-FGSM}), was introduced \cite{Dong_2018_CVPR}, which significantly enhanced the transferability of untargeted attacks on ImageNet dataset \cite{ILSVRC15} under a $l_\infty$ norm budget. More recently, \cite{xie2018improving} proposed a data augmentation technique named input diversity method (\textbf{DIM}) to further boost the transferability of these attack methods. In contrast to our self-supervised attack approach, all of these methods are supervised adversarial attacks that rely on cross-entropy loss to find the deceptive gradient direction.

\begin{figure*}[htp]
    \centering
    \includegraphics[width=1\textwidth]{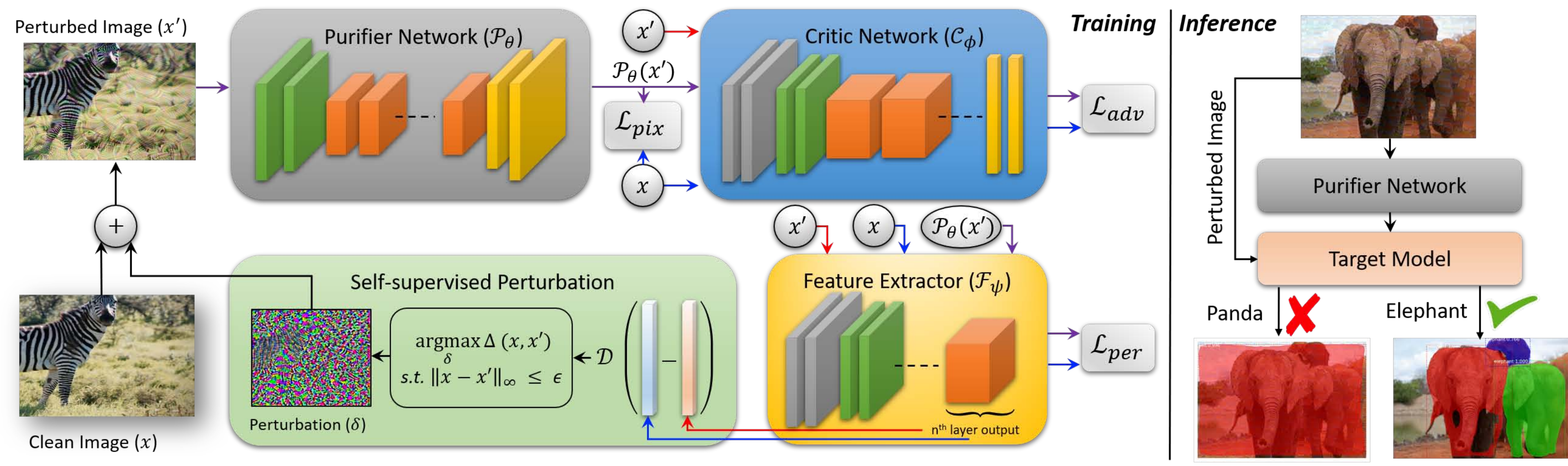}\vspace{-0.5em}
    \caption{\small \emph{Neural Representation Purifier}. Using a self-supervision signal, the proposed defense  learns to purify perturbed images, such that their corresponding perceptual representation in deep feature space becomes    close to clean natural images.}
    \label{fig:defense_overview}
\end{figure*}

%------------------------------------------------------------------------
%########## Overview of NRP
\section{Neural Representation Purifier} \label{sec:NRP}
Our defense aims to combine the benefits of adversarial training and input processing methods in a single framework that is computationally efficient, generalizable across different tasks and retains the clean image accuracy. The basic intuition behind our defense mechanism is to effectively use information contained in the feature space of deep networks to obtain an automatic supervisory signal. To this end, we design a Neural Representation Purifier (NRP) model that learns to clean adversarially perturbed images based on the automatically derived (self) supervision. 

The objective is to recover the original benign image $\bm{x}$ given an input adversarial image $\bm{x}'$. We wish to remove the adversarial patterns by training a neural network $\mathcal{P}_{\theta}$ parameterized by $\bm{\theta}$, which we refer as the purifier network. The main objective is to be independent of the task-specific objective function, such that once trained, the proposed defense is transferable to other models (even across tasks). Towards this end, the network $\mathcal{P}_{\theta}$ is trained in an adversarial manner by playing a game with the critic network $\mathcal{C}_{\phi}$, and a feature extractor $\mathcal{F}_{\psi}$ (see Fig.~\ref{fig:defense_overview}). The function of the purifier and critic networks is similar to generator and discriminator in a traditional Generative Adversarial Network (GAN) framework, with the key difference that in our case, $\mathcal{P}_{\theta}$ performs image restoration instead of image generation. The feature extractor, $\mathcal{F}_{\psi}$, is pretrained on ImageNet and remains fixed, while the other two networks are optimized during training. Adversarial examples $\bm{x}'$ are created by maximizing the $\mathcal{F}_{\psi}$'s response in random directions defined by a distance measure
(Algorithm~\ref{alg:SSP}), while at minimization step, $\mathcal{P}_{\theta}$ tries to recover the original sample $\bm{x}$ by minimizing the same distance (Algorithm~\ref{alg:NRP}).

%########## Self Supervision
\subsection{Self-Supervision}\label{subsec:self-supervision}
The automatic supervision signal to train NRP defense is obtained via a loss-agnostic attack approach.
%that directly feeds perturbations to our proposed NRP defense. 
Below, we first outline why such a Self-Supervised Perturbation (SSP) is needed and then describe our approach.

\noindent \textbf{Motivation:} Strong white-box attacks \cite{Kurakin2016AdversarialEI, carlini2017towards}, that are generally used for AT, consider already-known network parameters $\bm{\theta}$ and perturb the inputs to create $\bm{x'}$, such that they are misclassified by the target model, \ie $\mathcal{T}(\bm{x'}; \bm{\theta})\neq y$. Since the perturbations are calculated using gradient directions specific to $\bm{\theta}$, the resulting perturbed images $\bm{x'}$ do not generalize well to other networks \cite{Dong_2018_CVPR, Su2018, Dong_2018_CVPR, xie2018improving, Zhou_2018_ECCV}.  
This dependency limits these attacks to a specific network and task. In contrast, our goal is to design a self-supervised perturbation mechanism that can generalize across networks and tasks, thus enabling a transferable defense approach. The self-supervised perturbation is based on the concept of `\emph{feature distortion}', introduced next.

%######################################
% I-FGSM Vs MI-FGSM (NRDM vs number of iterations)
\begin{figure}[!tp]
  \centering
    \includegraphics[ width=0.9\linewidth]{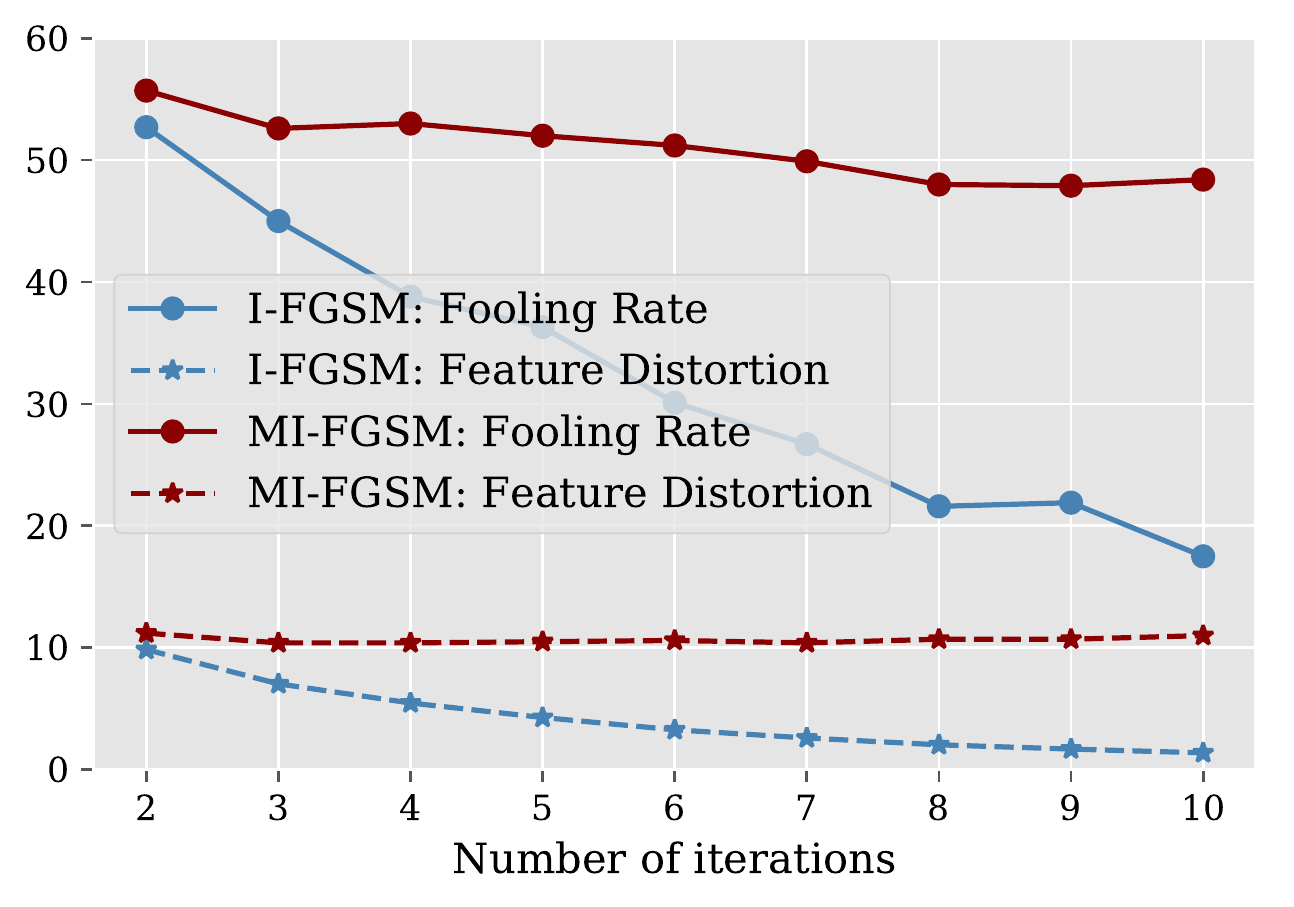} \vspace{-1em}
    % lelf lower right up  trim= 7mm 2mm 14mm 12mm, clip,
    \caption{\small Fooling rate of Inc-v4 and average feature distortion is shown for adversaries generated on Inc-v3 (black-box setting) by I-FGSM and MI-FGSM. 
    %Distortion is averaged over all examples. 
    As the number of iterations increases, fooling rate of I-FGSM decreases along with its feature distortion while MI-FGSM maintains its distortion as iterations increase.
    }
    \label{fig:iterations}
\end{figure}
%##############################################

\noindent  \textbf{Feature Distortion:}  Given a clean image $\bm x$ and its perturbed  counterpart $\bm{x'}$ that is crafted to fool the target model $\mathcal{T}(\cdot)$, the feature distortion refers to the change that $\bm{x'}$ causes to the internal representations of a neural network $\mathcal{F}(\cdot)$ relative to $\bm{x}$. This can be represented by,
\begin{align}\label{eq:distance_function}
      \Delta (\bm{x},\bm{x'}) = \bm{d}\big(\mathcal{F}(\bm{x}; \bm{\theta})|_n, \mathcal{F}(\bm{x'}; \bm{\theta})|_n \big),
\end{align}
where, $\mathcal{F}(\bm{x}; \bm{\theta})|_n$ denotes the internal representation obtained from the $n^{th}$ layer of a pretrained deep network $\mathcal{F}(\cdot)$ and $\bm{d}(\cdot)$ is a distance metric which can be $\ell_{p}$ \cite{goodfellow2014explaining}, Wasserstein distance \cite{arjovsky2017wasserstein} or cosine similarity between the features of the original and perturbed sample.

The reason why we base our self-supervised perturbation on feature distortion is its direct impact on the perturbation transferability.  To show this, we conduct a proof-of-concept experiment by  generating adversarial examples on ImageNet-NeurIPS \cite{Nips-data}. We consider two popular attack methods, MI-FGSM \cite{Dong_2018_CVPR} and I-FGSM \cite{Kurakin2016AdversarialEI}, among which MI-FGSM  has higher transferability compared to I-FGSM.  Interestingly, feature distortion strength of I-FGSM decreases as the number of attack iterations increases, compared to MI-FGSM (Fig.~\ref{fig:iterations}). MI-FGSM maintains its perturbation strength with increasing number of iterations. This indicates that feature distortion has a direct impact on  transferability and therefore maximizing the objective in Eq.~\ref{eq:distance_function} (signifying feature-space distortion) can boost the transferability of adversarial examples without using any decision boundary information. Based on this observation, our proposed perturbation generation approach directly maximizes the distortion in deep feature space to create strong, highly generalizable and task-independent adversarial examples.

%########## SSP: Self-Supervised Attack algorithm
\begin{algorithm}[t]
\small
\caption{SSP: Self-Supervised Perturbation}
\label{alg:SSP}
\begin{algorithmic}[1]
\Require A feature extractor $\mathcal{F}_{\psi}$, batch of clean samples $\bm{x}$, input transformation $\mathcal{R}$, perturbation budget $\epsilon$, step-size $\kappa$, and number of iterations $T$.
\Ensure Perturbed sample $\bm{x'}$ with $\|\bm{x'} - \bm{x}\|_{\infty} \leq \epsilon$.
\State $\bm{g}_0 = 0$; $\bm{x'} = \mathcal{R}(\bm{x})$;
\For {$t = 1$ to $T$}
\State Forward pass $\bm{x'}_t$ to $\mathcal{F}_{\psi}$ and compute $\Delta$ using Eq.~\ref{eq:distance_function};
\State Compute gradients $\bm{g}_{t} = \nabla_{\bm{x}} \,\Delta (\bm{x}_t, \bm{x'})$;
\State Generate adversaries using;
\vspace{-2ex}
\begin{equation}
\bm{x'}_{t+1} = \bm{x'}_{t} +\kappa\cdot\mathrm{sign}(\bm{g}_{t});
\end{equation}
\vspace{-4ex}
\State Project adversaries in the vicinity of $\bm{x}$
\vspace{-2ex}
\begin{equation}
\bm{x'}_{t+1} = \mathrm{clip}(\bm{x'}_{t+1}, \hspace{1ex} \bm{x}-\epsilon,  \hspace{1ex} \bm{x}+\epsilon);
\end{equation}
\vspace{-4ex}
\EndFor \\
\Return $\bm{x'} = \bm{x'}_T$.
\end{algorithmic}
\end{algorithm}

\noindent \textbf{Self-supervised Perturbation:} Conventional black-box attacks operate in the logit-space of deep networks. The objective of `\textit{logit-based}' adversarial attacks is to change the target model's prediction for a clean image $\mathcal{T}(\bm{x}) \neq \mathcal{T}(\bm{x}')$ such that $\bm{x'}$ is bounded: ${\parallel} \bm{x} - \bm{x'} {\parallel} \le \epsilon$. 
In contrast to  these methods, we propose to find adversaries by maximizing the feature loss (Sec.~\ref{NRP Loss functions}) of neural networks.  Our approach does not rely on decision-boundary information since our `\textit{representation-based}' attack directly perturbs the feature space by solving the following optimization problem:
\begin{align}\label{eq:NRD}
 \underset{\bm{x}'}{\text{max}}  \;\; \Delta(\bm{x},\bm{x'}) \; \,
 \text{subject to:}  \; {\parallel} \bm{x} - \bm{x}' {\parallel}_{\infty} \leq \epsilon,
\end{align}
 
Our proposed method to maximize feature distortion for a given input sample is summarized in Algorithm \ref{alg:SSP}.
We apply a transformation $\mathcal{R}$ to input $\bm{x}$ at the first iteration (Algorithm \ref{alg:SSP}) to create a neural representation difference between an adversarial and benign example and then maximize the difference within a given perturbation budget. There can be different choices for $\mathcal{R}$ but in this work, $\mathcal{R}$ simply adds random noise to the input sample, \ie our algorithm takes a random step at the first iteration.

%##########  NRP: algorithm
\begin{algorithm}[t]
\small
\caption{NRP: Neural Representation Purification via Self-Supervised Adversarial Training}
\label{alg:NRP}
\begin{algorithmic}[1]

\Require Training data  $\mathcal{D}$, Purifier $\mathcal{P}_{\theta}$, feature extractor $\mathcal{F}_{\psi}$, critic network $\mathcal{C}_{\phi}$, perturbation budget $\epsilon$ and loss criteria $\mathcal{L}$.

\Ensure Randomly initialize $\mathcal{P}_{\theta}$ and $\mathcal{C}_{\phi}$.
\Repeat
\State Sample mini-batch of data, $\bm{x}$, from the training set.
\State Find adversaries, $\bm{x}'$, at a given perturbation budget $\epsilon$ by 
\Statex \hspace{0.4cm} maximizing distance, $\Delta$ (Eq. \ref{eq:distance_function}), using Algorithm \ref{alg:SSP}.
\State Forward-pass $\bm{x}'$ through $\mathcal{P}_{\theta}$ and calculate $\mathcal{L}_{\mathcal{P}_{\theta}}$  (Eq. \ref{eq:geneator_overlall_loss}).
\State Back-pass and update  $\theta$ to minimize 
$\mathcal{L}_{\mathcal{P}_{\theta}}$  (Eq. \ref{eq:geneator_overlall_loss}).
\State Update $\mathcal{C}_{\phi}$ to classify $\bm{x}$ from $\mathcal{P}_{\theta}(\bm{x}')$.
\Until{ $\mathcal{P}_{\theta}$ converges.}
\end{algorithmic}
\end{algorithm}

%########## Losses
\subsection{NRP Loss functions}\label{NRP Loss functions}
We propose a hybrid loss function that is used to train the purifier network (see Algorithm~\ref{alg:NRP}). This loss function consists of three terms that we explain below:\\
\textbf{Feature loss:} The Self-supervised Perturbation (SSP) generated by Algorithm~\ref{alg:SSP} is the direct result of increasing the feature loss function, $\Delta$, defined on the feature extractor $\mathcal{F}_{\psi}$. In order to learn the purifier network, we must decrease this distance as follows: 
\begin{align}\label{eq:feature_loss}
	\mathcal{L}_{feat} = \Delta \big(\,\F(\bm{x}), \; \F(\mathcal{P}_{\theta}(\bm{x}'))\,\big),
\end{align}
where, $\Delta$ is formally defined in Eq.~\ref{eq:distance_function},  and the distance measure used to compute $\Delta$ is the mean absolute error (MAE). We empirically observe that removing $\mathcal{L}_{feat}$ loss leads to a network that does not converge to a meaningful state  and produces weaker defense (see Fig. \ref{fig:ablation_nrp}).% This indicates the importance of $\mathcal{L}_{feat}$ in removing adversarial noise. 

\noindent
\textbf{Pixel loss:} Smoothing images can help in mitigating the adversarial effect since the perturbation patterns resemble to that of noise. Therefore, in order to encourage smoothness, we apply $l_2$ loss in the image pixel space, 
%that is computed using $\mathcal{P}_{\theta}(\bm{x}')$ and $\bm{x}$:
\begin{align}
	\mathcal{L}_{img} = \|\mathcal{P}_{\theta}(\bm{x}') - \bm{x} \|_{2}.
\end{align}

\noindent
\textbf{Adversarial loss:} Instead of using vanilla GAN objective, we use relativistic average GAN which has shown better convergence properties \cite{jolicoeur2018relativistic,naseer2019cross}. For a given batch of original, $\bm{x}$, and adversarial examples, $\bm{x}'$, the relativistic loss for the purifier network $\mathcal{P}_{\theta}$ is given as:
 \begin{align}
    %  \mathcal{L}_{adv} = - \frac{1}{N} \sum_{i=1}^{N} \log\big(\, \mathcal{C}_{\phi}(\mathcal{P}_{\theta}(\bm{x}')) - \mathcal{C}_{\phi}(\bm{x})\,\big),
      \mathcal{L}_{adv} = - \log\big(\, \sigma\big(\, \mathcal{C}_{\phi}(\mathcal{P}_{\theta}(\bm{x}')) - \mathcal{C}_{\phi}(\bm{x})\,\big)\,\big),
 \end{align}
 where $\sigma$ represents the sigmoid layer. The overall loss objective for $\mathcal{P}_{\theta}$  is the combination of losses defined on pixel and feature spaces as well as the relativistic loss:
\begin{equation}\label{eq:geneator_overlall_loss}
    \mathcal{L}_{\mathcal{P}_{\theta}} =
    \underbrace{\alpha\cdot\mathcal{L}_{adv}}_{Adversarial\:loss}  + \underbrace{\gamma\cdot\mathcal{L}_{img}}_{Pixel\:loss}+  \underbrace{\lambda \cdot\mathcal{L}_{feat}}_{Feature \:loss} .
\end{equation}
The pixel and feature losses focus on restoring image content and style, while adversarial loss restores texture details.

%########## Architectures
\subsection{NRP Architecture}
Here, we outline the architecture of generator, feature extractor and discriminator blocks.  \textbf{Generator ($\mathcal{P}_{\theta}$):} Our generator architecture is inspired by \cite{ledig2017photo, wang2019edvr}. It consists of a convolution layer followed by multiple ``basic blocks". Each basic block is composed of 3 ``dense blocks" and each dense block contains five convolutional layers followed by leaky-relu \cite{xu2015empirical} and finally a convolutional layer that has output with same dimension as input. Generally, adding a skip connection from input to generator's output helps in restoration tasks e.g., image super resolution \cite{ledig2017photo} and deblurring \cite{Kupyn_2018_CVPR}. However, in our case an important design criteria is to avoid such skip connection since our objective is to remove adversarial noise and a direct skip connection can potentially reintroduce harmful noise patterns.
\textbf{Feature Extractor ($\mathcal{F}_{\psi}$):} It is a VGG \cite{Simonyan14verydeep} network pretrained on ImageNet. During training, $\mathcal{F}_{\psi}$ remains fixed while its response is maximized in random directions (adversary generation process) and minimized (purification process) using a predefined distance metric. In our experiments, we demonstrate the effectiveness of VGG space for creating strong adversaries as compared to other deep architectures. 
\textbf{Discriminator ($\mathcal{C}_{\phi}$):} Our discriminator architecture is also based on VGG network \cite{Simonyan14verydeep}. It consists of five convolutional blocks containing convolutional layers followed by batch-norm and leaky-relu and then a fully connected layer.

\begin{figure}[t]
\centering
  \begin{minipage}{.23\textwidth}
  	\centering
  	 % lelf lower right up 
    \includegraphics[  trim= 7.5mm 0mm 0mm 10mm, width=\linewidth, keepaspectratio]{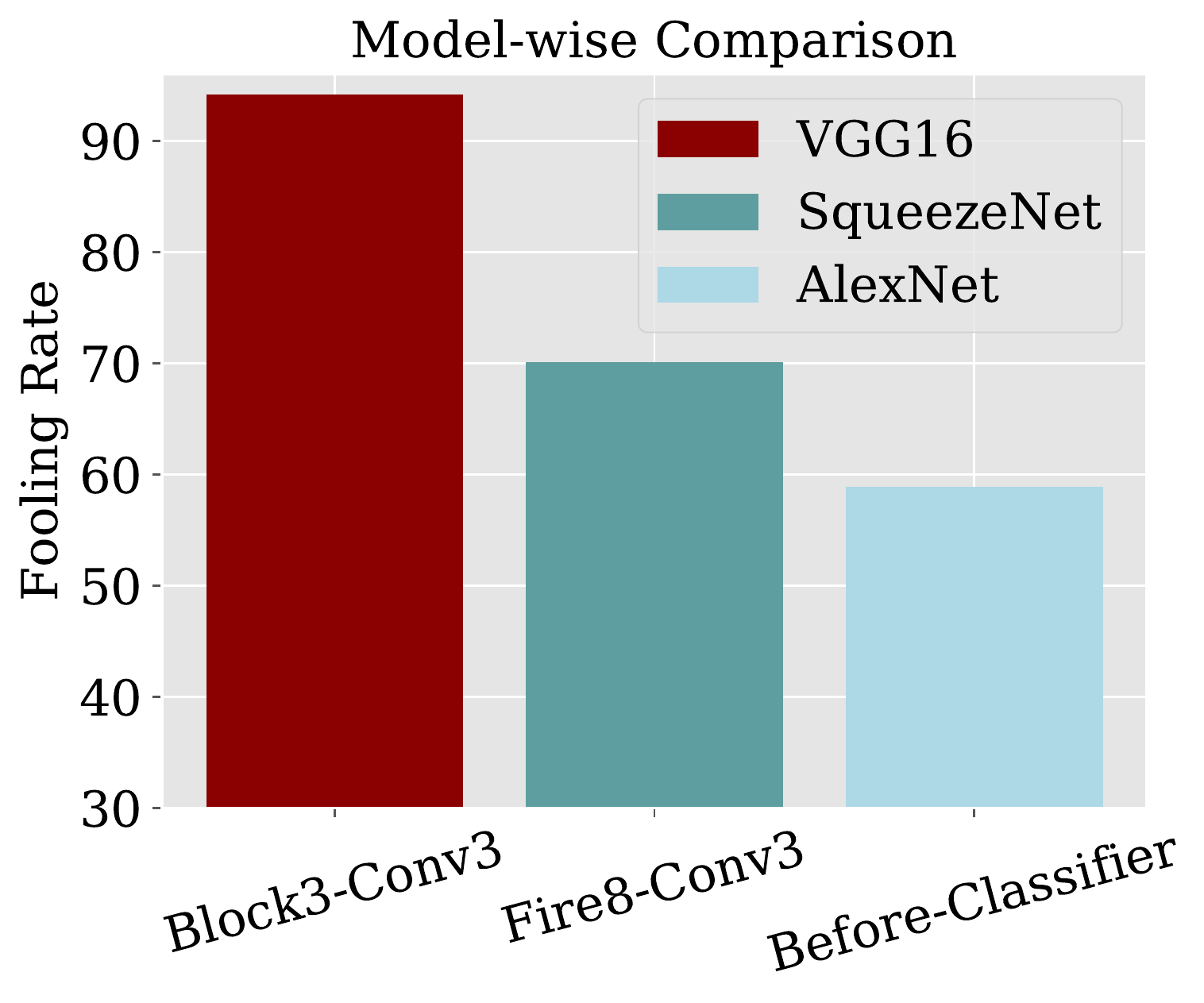}
  \end{minipage} 
  \begin{minipage}{.23\textwidth}
  	\centering
  	 % lelf lower right up 0.
  	 
    \includegraphics[  trim= 7.5mm 0mm 0mm 10mm,width=\linewidth, keepaspectratio]{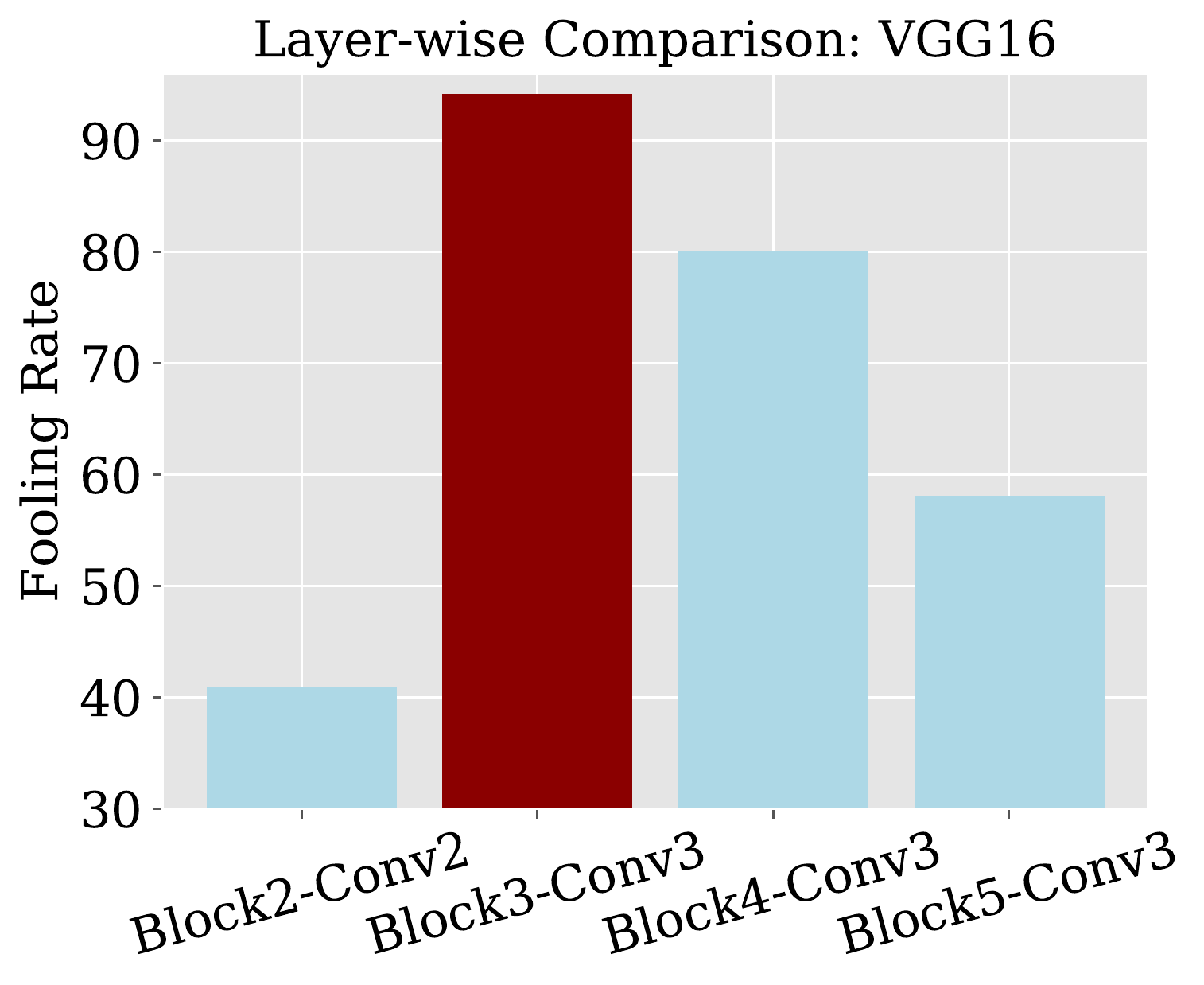}
  \end{minipage} \vspace{-2ex}
  \caption{\small Fooling rate for Inc-v3 \cite{kurakin2016adversarial} on ImageNet-NeurIPS dataset. Adversaries are created by applying SSP (Algorithm \ref{alg:SSP}) at different layers and best results for each model is selected. Perceptual adversaries found in VGG space has the highest transferability (further analysis is in supplementary material).}
\label{fig:alex-vs-vgg16-vs-sq}
\end{figure}

%######## Perceptual Adversaries
\subsection{On Suitable Perceptual Adversaries} 
The intuition to train NRP on boundary-agnostic perceptual adversaries is based on the extensive study \cite{Zhang2018TheUE} that found correlation of deep features with human perception. Specifically, \cite{Zhang2018TheUE} compares three models \ie VGG \cite{Simonyan14verydeep}, AlexNet \cite{Krizhevsky2012ImageNetCW} and SqueezeNet \cite{Iandola2017SqueezeNetAA}. Following \cite{Zhang2018TheUE}, we study these models from adversarial perspective by applying feature distortion at different layers in Fig.~\ref{fig:alex-vs-vgg16-vs-sq}. Our findings are as follows: \textbf{(a)} VGG's perceptual adversaries are more transferable than AlexNet and SqueezeNet (a detailed transferability analysis on seen/unseen perturbations of VGG is in supplementary material), \textbf{(b)} under same feature distortion settings, adversaries found at different layers are not equally transferable \eg  conv3.3 (block 3, layer 3) features offer better adversarial transferability than the rest of the network. We believe this is because the initial VGG layers learn low-level features while the deeper ones become too specific to the label space. Further, we found that increasing the representation loss at multiple network layers does not notably increase attack success rate and adds a significant computational overhead. 
Since NRP training process is agnostic to the label-space of the source model i.e., it neither depends on a particular task-specific loss function (e.g., cross entropy) nor on the ground-truth labels, this makes it a generic algorithm, which can defend a totally unseen model. Furthermore, we demonstrate that perturbations discovered with our SSP approach offer high transferability across models trained on different datasets and tasks.

%-------------------------------------------------------------------------
\section{Experiments}

\subsection{Training Details}\label{subsec: training_details}
Training is done on randomly selected 25k images from MS-COCO data set. These images are resized to $480\times480\times3$. Adversaries created using SSP are fed as inputs to NRP with their corresponding clean images used as target labels. During training, we randomly crop images of $128\times128\times3$. Batch size is set to 16 and training is done on four Tesla v100 GPUs. Learning rates for generator and discriminator are set to $10^{-4}$, with the value of $\alpha = 5\times10^{-3}$,  $\gamma = 1\times10^{-2}$ and $\lambda=1$.
We study eight models trained on the ImageNet \cite{ILSVRC15}. Five of these models are naturally trained. These include Inceptionv3 (Inc-v3) \cite{szegedy2016rethinking}, Inceptionv4 (Inc-v4),  Inception Resnet v2 (IncRes-v2)  \cite{szegedy2017inception},  Resnet v2-152 (Res-152) \cite{he2016identity} and VGG-19 \cite{Simonyan14verydeep}. The other three  models including Adv-v3 \cite{kurakin2016adversarial}, Inc-v3\textsubscript{ens3} and IncRes-v2\textsubscript{ens} \cite{tramer2017ensemble} are adversarially trained. The specific details about these models can be found in \cite{kurakin2016adversarial, tramer2017ensemble}.

\input{tab_def_1.tex}

\input{fig_tab_cross_defense.tex}

\subsection{Defense Results and Insights}\label{sec:def_res}
%\mz{I couldn't write this section well. Need a little bit of discussion on to summarize all the results well.}
\noindent \textbf{(a) Generalizability Across Attacks:} Figs.~\ref{fig:NRP-removing-perturbation},~\ref{fig:NRP-classification-demo}~\&~\ref{fig:NRP-detection-demo} demonstrate generalization ability of NRP to recover images from strong adversarial noise. Quantitative analysis in Table~\ref{tab:NRP-main-defense-results} shows that compared to previously broken defenses \cite{dong2019evading}, NRP achieves strong robustness against state-of-the-art attacks \cite{xie2018improving, dong2019evading}, bringing down the effectiveness of the \textbf{ensemble} translation-invariant attack with input diversity (DIM$_{TI}$) \cite{dong2019evading} from $79.8\%$  to $31.9\%$.

\begin{figure}[h]
\centering
  \begin{minipage}{.48\textwidth}
  	\centering
  	 % lelf lower right up 
    \includegraphics[ trim= 7mm 2mm 0mm 5mm, width=0.95\linewidth, keepaspectratio]{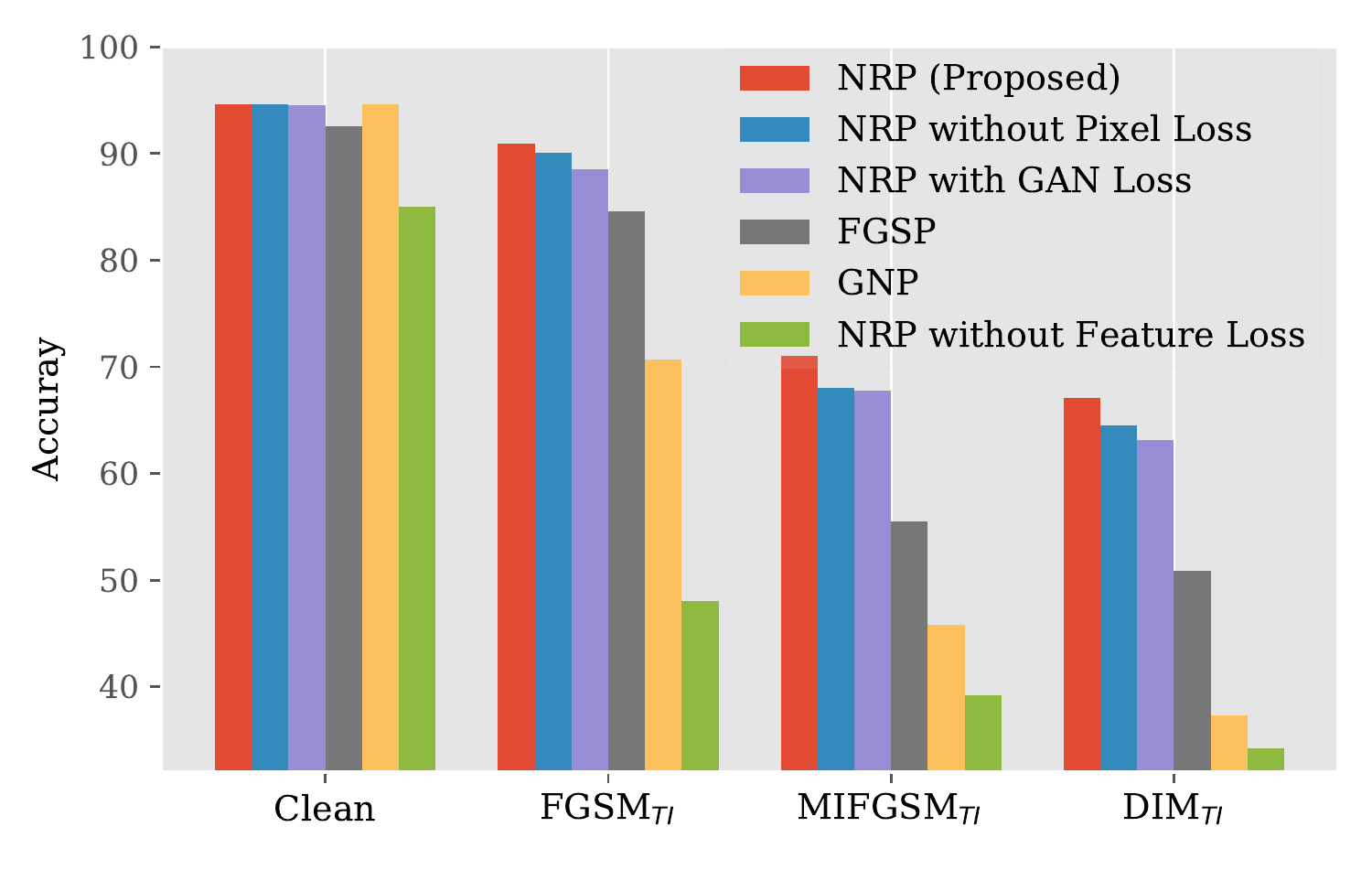}\
  \end{minipage}
  \vspace{-1em}
  \caption{\small Ablation. %purifier networks trained via our approach. 
  Proposed NRP is able to recover input samples from the strong black-box \textbf{ensemble} attack \cite{dong2019evading} as compared to GNP %(trained to purify Gaussian-noise samples) 
  and FGSP. %(trained to purify adversaries found using FGSM). 
  NRP trained without $\mathcal{L}_{feat}$ performs poorly indicating the importance of perceptual loss. Top-1 accuracy (\emph{higher is better}) is reported for IncRes-v2\textsubscript{ens} \cite{tramer2017ensemble} on ImageNet-NeurIPS.}\vspace{-1em}
\label{fig:ablation_nrp}
\end{figure}
%###############################################################

\input{fig_tab_classification.tex} %3 vis figs; what iff attacker knows

\noindent \textbf{(b) NRP as Cross-task Defense:} In order to measure the cross-task defense capabilities, we deploy NRP against cross-domain attack (CDA) \cite{naseer2019cross}, a state-of-the-art attack that generates diverse cross-domain adversarial perturbations. Results in Table \ref{tab:NRP-Classfication-Segmentation-Detection} demonstrate that NRP successfully removes all unseen perturbations and proves a generic cross-task defense for classification, object detection and instance level segmentation against CDA.

\noindent \textbf{(c) Ablation:} Fig.~\ref{fig:ablation_nrp} thoroughly investigates the impact of different training mechanisms in combination with our defense, and provides the following insights:
\textbf{(i)} Relativistic GAN loss offers a more robust solution than vanilla GAN,
\textbf{(ii)} NRP performance decreases slightly without pixel loss,
 \textbf{(iii)} NRP without feature loss loses supervisory signal defined by perceptual-space boundary, hence the generator does not converge to a meaningful state,
\textbf{(iv)} Gaussian smoothing (Gaussian noise data augmentation) proves to be useful in reducing adversarial vulnerability of classifier \cite{cohen2019certified, Zantedeschi2017EfficientDA}. Training NRP as a Gaussian denoiser, named Gaussian Noise Purifier (GNP) does not prove effective against translation-invariant attacks \cite{dong2019evading}, and
\textbf{(v)} Training NRP to stabilize FGSM adversaries (termed FGSP in Fig.~\ref{fig:ablation_nrp}) performs relatively better than GNP.

%####################################################

\noindent \textbf{(d) What if Attacker knows about the Defense:} We study this difficult scenario with the following criteria: \textbf{(i)} attacker knows that the defense is deployed and has access to its training data and training mechanism, and \textbf{(ii)} attacker trains a local defense similar to NRP, and then uses BPDA \cite{carlini2017towards} to bypass the defense. To simulate this attack, we train residual generator (ResG) \cite{ledig2017photo} and UNet \cite{ronneberger2015u} with the same training mechanise as described in Sec. \ref{subsec: training_details}. We then combine BPDA \cite{athalye2018obfuscated} with translation-invariant attack to bypass NRP. Under these challenging settings, NRP shows a relative gain of $74\%$ and $66\%$ respectively for IncRes-v2, IncRes-v2\textsubscript{ens} (see Table~\ref{tab:bpda_demo}).

\subsection{Self Supervised Perturbation as an Attack}\label{subsec:self-supervised-attack}

\input{ssp_attack_tables.tex}

Next, we evaluate the strength of SSP as an attack for the tasks of classification, detection and segmentation.

\noindent \textbf{Classification:} Table \ref{tab:att-nips} compares SSP with FGSM \cite{goodfellow2014explaining}, R-FGSM \cite{tramer2017ensemble}, I-FGSM \cite{Kurakin2016AdversarialEI}, MI-FGSM \cite{Dong_2018_CVPR}, TAP \cite{Zhou_2018_ECCV} and DIM \cite{xie2018improving} using their standard hyper-parameters (see supplementary material). The results in Table~\ref{tab:att-nips} provide the following insights. \textbf{(i)} SSP consistently demonstrates a strong black-box adversarial transferability on both naturally and adversarially trained models, bringing down top-1 accuracy of IncRes-v2 \cite{szegedy2017inception} from $100.0\%$ to $14.1\%$, \textbf{(ii)} While MI-FGSM \cite{Dong_2018_CVPR} and DIM \cite{xie2018improving} perform slightly better on adversarially trained ensemble models \cite{tramer2017ensemble} in terms of top-1 accuracy, SSP shows comparable top-1 rate and surpasses in terms of top-5 accuracy, and \textbf{(iii)} These results indicate that decision-boundary based attacks flip the label of input sample to the near-by class category, while SSP being agnostic to decision-level information pushes the adversaries far from the original input category. 
%#######################################################################################
\noindent \textbf{Cross-task Adversarial Attack:}
Since SSP is loss-agnostic, it enables attacks on altogether different tasks. Table~\ref{tab:SSP-Segmentation-Detection} explores SSP for object detection and image segmentation. 
For Segmentation, the self-supervised perturbations created on CAMVID \cite{brostow2009semantic} in VGG-16 feature space are able to bring down the per pixel accuracy of Segnet-Basic by $47.11\%$ within $l_\infty \le 16$. For object detection, on MS-COCO validation set \cite{lin2014microsoft}, mean Average Precision (mAP) with $0.5$ intersection over union (IOU) of RetinaNet \cite{lin2018focal} and Mask-RCNN \cite{He2017MaskR} drop from $53.78\%$ to $5.16\%$  and $59.5\%$ to $9.7\%$, respectively, under $l_\infty \le 16$.

\section{Conclusion}\vspace{-0.5em}
We propose a novel defense approach that removes harmful perturbations using an adversarially trained purifier. Our defense does not require large training data and  is independent of the label-space.  It exhibits a high generalizability to the unseen state-of-the-art attacks and successfully defends a variety of tasks including classification, segmentation and object detection. Notably, our defense is able to remove structured noise patterns where an adversarial image is maliciously embedded into the original image.

{\small
\bibliographystyle{ieee_fullname}
\bibliography{egbib}
}

\clearpage
\setcounter{section}{0}
\def\thesection{Appendix \Alph{section}}
\setcounter{table}{0}
\setcounter{figure}{0}
\section*{Supplementary: A Self-supervised Approach for Adversarial Robustness}
We first explore why Self-supervised Perturbation (SSP) attack works in \ref{sec:why-SSP-works}. In \ref{sec:comparison_with_feature_denoising}, we compare NRP with conventional adversarial training (AT) method known as feature denoising \cite{xie2019feature} in terms of adversarial robustness and defense training time. Differences of our proposed attack and defense from feature scattering \cite{zhang2019defense} method are discussed in \ref{sec:fs}. Ability of SSP to fool object detectors is compared against CDA \cite{naseer2019cross} in \ref{sec:ssp_vs_cda}. 
We show that different transformation based defenses, JPEG, total variation minimization (TVM) and median filtering (MF) are not effective against SSP in \ref{subsec:input-transformation-SSP}. Attack parameters against which our defense is evaluated are provided in \ref{subsec:attack-parameters}.  Finally, we visually demonstrate NRP's ability to remove different kinds of adversarial perturbations in \ref{subsec:NRP-genralization-demo}.

\section{Why Self-supervision Works?}\label{sec:why-SSP-works}
Here, we highlight our intuition to create adversarial examples using feature space of VGG model \cite{Simonyan14verydeep}.

\begin{itemize}[leftmargin=*]
    \item \textbf{Neural Style Transfer:} \cite{engstrom2019a, nakano2019robuststyle} observed that the ability to transfer styles improves with AT, a phenomenon often related to VGG models \cite{Simonyan14verydeep} . On the other hand, VGG networks are more vulnerable to adversarial attacks \cite{engstrom2019a}. A hypothesis was presented in \cite{engstrom2019a} that perhaps VGG initial layers are as robust as adversarially trained models which allows better style transfer without AT. 
    
%#######################################################
\begin{figure}[t]
\centering
  \begin{minipage}{.48\textwidth}
  	\centering
  	 % left lower right up 
    \includegraphics[ trim= 7mm 2mm 0mm 5mm, width=\linewidth, keepaspectratio]{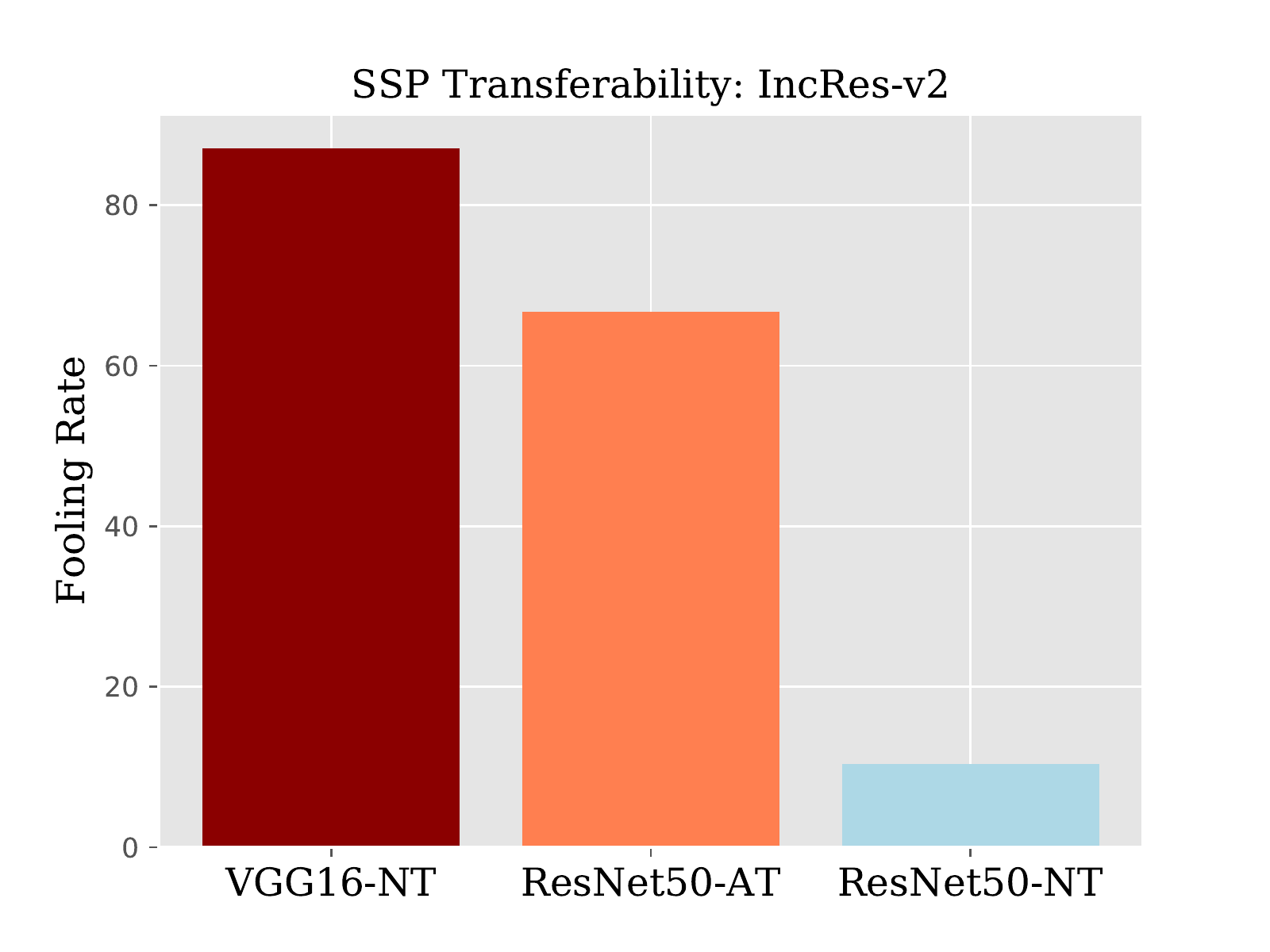}\
  \end{minipage}
  \caption{Fooling rate comparison is presented. NT and AT represent naturally and adversarially trained models, respectively. VGG16-NT and ResNet50-NT are trained on ImageNet while ResNet50-AT \cite{engstrom2019adversarial} is adversarially trained on a subset of ImageNet. Adversaries are created by applying distortion to the feature space of each model on NeurIPS dataset and then transferred to naturally trained IncRes-v2 \cite{szegedy2017inception}. Adversaries found in VGG space have higher transferability. In comparison, transferability of feature space of ResNet50 increases after adversarially training.} 
\label{fig:transferability_vgg16_vs_res50_vs_robust_res50}
\end{figure}
%#######################################################  

%#######################################################
\begin{figure*}[t]
\centering
  \begin{minipage}{.24\textwidth}
  	\centering
  \includegraphics[width=\linewidth,keepaspectratio]{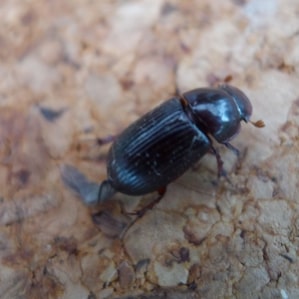}\
  Clean Image
  \end{minipage}
    \begin{minipage}{.24\textwidth}
  	\centering
  \includegraphics[width=\linewidth, keepaspectratio]{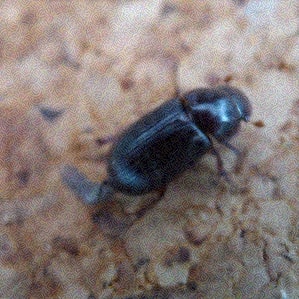}\
  Adversarial: ResNet50-NT 
  \end{minipage}
    \begin{minipage}{.24\textwidth}
  	\centering
  \includegraphics[width=\linewidth, keepaspectratio]{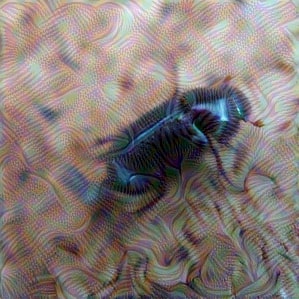}\
  Adversarial: VGG16-NT
  \end{minipage}
    \begin{minipage}{.24\textwidth}
  	\centering
   \includegraphics[width=\linewidth, keepaspectratio]{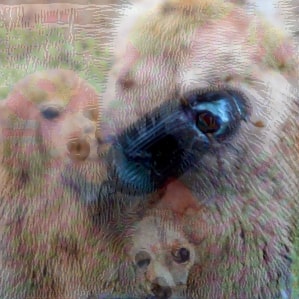}\
   Adversarial: ResNet50-AT
  \end{minipage}
  \caption{A visual demonstration of adversaries found by SSP in the feature space of diffrent networks. Perturbation buget is set to $l_\infty \le 16$. NT and AT represent naturally and adversarially trained models, respectively. }
  \label{fig:demo_vgg16_vs_res50_vs_robust_res50}
\end{figure*}
%#########################################################
\begin{figure*}[!htp]
\centering
  \begin{minipage}{.48\textwidth}
  	\centering
  	 % lelf lower right up 
    \includegraphics[ width=\linewidth, keepaspectratio]{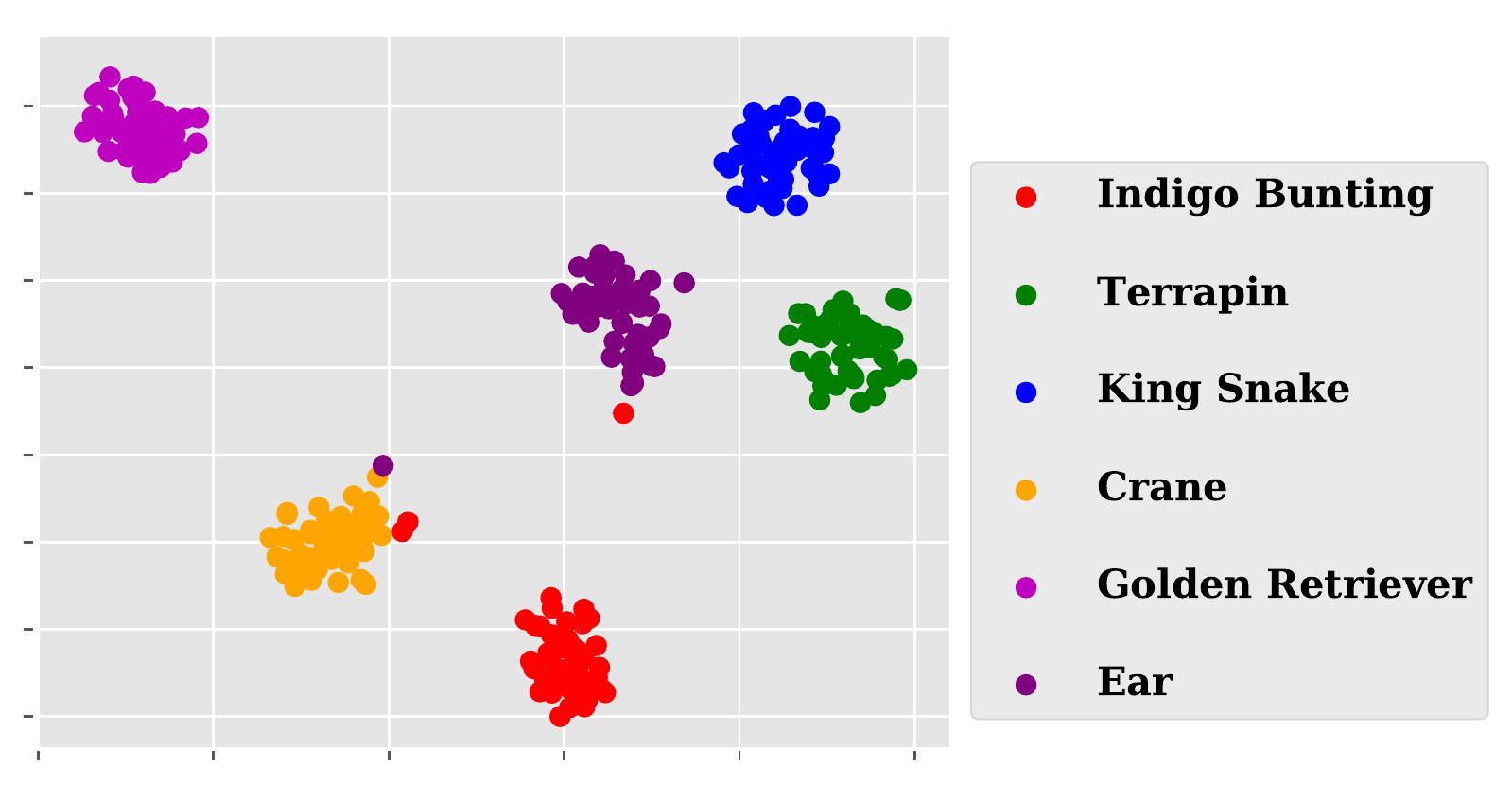}\
     \small (b) Logits 
  \end{minipage}
    \begin{minipage}{.48\textwidth}
  	\centering
  	 % lelf lower right up 
    \includegraphics[width=\linewidth, keepaspectratio]{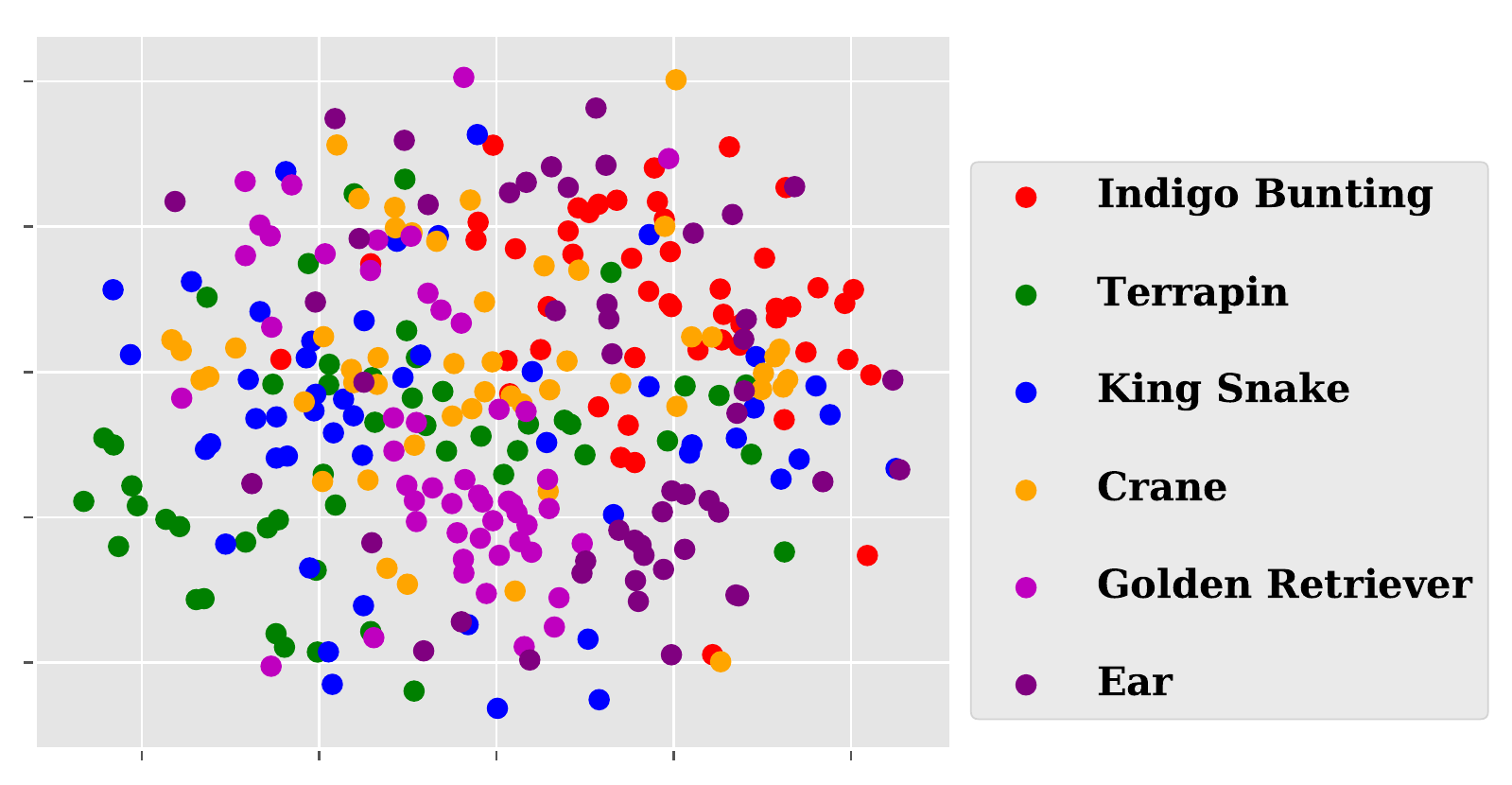}\
    \small (b) Features 
  \end{minipage}
  \caption{\small t-SNE \cite{maaten2008visualizing} visualization of logits vs. feature representation of randomly selected classes from ImageNet validation set. Logits are computed from VGG16 \cite{Simonyan14verydeep} last layer while features are extracted from "Block3-Conv3" of the same model. Our intuition is based on the observation that features space is shared among input samples rather than the logit space. Attacking such shared representation space removes task dependency constraint during adversary generation optimization and produces generalizable adversarial examples.}
\label{fig:tsne_logits_vs_features}
\end{figure*}
%###################################################

    \item \textbf{Transferability of Natural vs. Robust Layers:} In addition to style transfer hypothesis \cite{engstrom2019a}, we explore the connection between layers of VGG and adversarially trained models in the context of adversarial attacks:
    \begin{itemize}[leftmargin=*]
        \item \textbf{Maximum Distortion of Non-Robust Features:} Datasets containing natural images contain both robust and non-robust features \cite{ilyas2019adversarial}. Robust features can be described by high level concepts like shape \eg ear or noise etc., while non-robust features can arise from background or texture \cite{geirhos2018imagenettrained}. Ilyas \etal \cite{ilyas2019adversarial} argues that neural networks can pick-up on non-robust features to minimize the empirical risk over the given the data distribution %($\mathbb{E}_{(\bm{x},y)\sim D}[J(\bm{x},y,\theta)]$) 
        and the transferability of adversarial examples can be explained by these non-robust features in different networks. 
        \item \textbf{Transferability:} VGG's ability to destroy non-robust features translates to better transferability even without any AT as compared to ResNet models (see Figures \ref{fig:transferability_vgg16_vs_res50_vs_robust_res50} and \ref{fig:demo_vgg16_vs_res50_vs_robust_res50}). 
    \end{itemize}

    \item \textbf{Shared Representation Space:} Our objective is to find adversarial patterns that can generalize across different network architectures trained for different tasks (\eg classification, objection detection or segmentation). These are diverse tasks that do not share loss functions, dataset or training mechanism. Decision-boundary based attacks use model final response (\eg logits in the case of classification) that is specific to input sample which leads to task-specific perturbations. A network's feature space, however, is shared regardless the input category. Therefore, perturbations found in such a space are highly generalizable (see Figure \ref{fig:tsne_logits_vs_features}).
\end{itemize}

\section{Comparison with AT}
\label{sec:comparison_with_feature_denoising}
Conventional AT methods, such as \cite{xie2019feature}, lose clean accuracy to gain adversarial robustness.  Take an example of ResNet152 adversarially trained by \cite{xie2019feature}. In order to gain 55.7\% robustness ($\epsilon \le 16$) against targeted PGD attacks with ten number of iterations, the model clean accuracy drops from 78\% to 65.3\% which is even lower than VGG11. In contrast, our approach does not suffer from performance degradation on clearn samples. 

\subsection{Defense Results}
To compare against \cite{xie2019feature}, we ran ten number of PGD attack iterations. Labels for this targeted attack were chosen randomly as suggested by \cite{xie2019feature}. It is important to note that NRP can be turned into a dynamic defense, for example by first taking a random step in the input space and then projecting the modified input sample onto the perceptual space using our NRP. This way, NRP can be used to defend against attacks that try to incorporate NRP during attack optimization (a white box setting). We demonstrate this behavior in Table \ref{tab:1} by incorporating NRP in PGD attack using backpass approach introduced in \cite{athalye2018obfuscated}. Even for this challenging scenario, NRP shows significantly higher robustness than \cite{xie2019feature} while maintaining a higher clean accuracy. This highlights the benefit of self-supervision in AT. 

\begin{table}[!h]
\setlength\tabcolsep{5.0pt}
    %\footnotesize
    \begin{center}
    \resizebox{0.8\columnwidth}{!}{
            \begin{tabular}{@{}l c c@{}}
            \toprule
            \multirow{2}{*}{Method} & \multirow{2}{*}{Clean} &\multicolumn{1}{c}{Adversarial}\\
            \cmidrule(lr){3-3} 
             &  & $\epsilon \le 16/255$ \\
            \midrule
            Original & \textbf{78.31} & 0.66\\
            Feature Denoising\cite{xie2019feature} &65.3 &55.7 \\
            \midrule
            NRP & 73.5 $\pm$ 1.5 & \textbf{63.0} $\pm$ \textbf{2.0}\\
            \bottomrule
            \end{tabular}
            }
             \end{center}
        \vspace{-0.3cm}
        \caption{Defense success in terms of accuracy on ImageNet validation set (50k images). Higher is better.}
        \label{tab:1}
\end{table}

\subsection{Training Cost}
Conventional AT methods like \cite{xie2019feature} depend on number of classes, dataset and task. In contrast, our defense is independent of such constraints. We describe the computational benefits of our defense with feature denoising based AT \cite{xie2019feature} in Table~\ref{tab:2}. Training time of our defense remains the same regardless of the backbone model while training time for \cite{xie2019feature} increases with the model size. In conclusion, conventional AT requires large amount of labelled data (\eg, \cite{xie2019feature} is trained on 1.3 million images of ImageNet), while our defense can be trained on small unlabelled data (\eg, 25k unlabelled MS-COCO images).  

\begin{table}[!h]
    \centering %\setlength{\tabcolsep}{4pt}
    \scalebox{0.9}{
        \begin{tabular}{cccccc}
            \toprule
            Method & No. of & Training & Task/Label  & Dataset \\
             &  GPUs &  Time  & Dependency & Specific\\
            \midrule
            \cite{xie2019feature} & 128 & 52 & Yes & Yes \\
            NRP& 4 & 28 & No & No \\
            \bottomrule
          \end{tabular}}
        \caption{  Comparison of training time (hours) between NRP and AT  on ResNet152 model \cite{xie2019feature}.}
        \label{tab:2}
\end{table}

\section{Comparison with \cite{zhang2019defense}}
\label{sec:fs}
\textit{Defense comparison:} Feature Scattering (FS) \cite{zhang2019defense} based AT remains model and task-specific. Instead, our defense is independent to the target model and task, thereby providing better generalizability. 

\textit{Attack comparison:}~The proposed attack FSA  \cite{zhang2019defense} operates in logit space in an unsupervised way by maximizing Optimal Transport distance, as compared to our SSP which operates in perceptual feature space (\eg, VGG features).  we compare the transferability of their attack with our SSP. As demonstrated in Table \ref{tab:3}, SSP performs favorably well against FSA. 

\begin{table}[!h]
    \centering
    \scalebox{0.7}{
        \begin{tabular}{ccccccc}
            \toprule
            Attack & Inc-v3 & Inc-v4 & Res-152 & IncRes-v2 & Adv-v3 & IncRes-v2\textsubscript{ens}\\
            \midrule
            FSA \cite{zhang2019defense} & 60.4&64.2&68.8&71.0&72.2&88.6\\
            SSP& \textbf{5.3} &\textbf{ 5.9} & \textbf{16.5} & \textbf{14.1} & \textbf{25.9} &\textbf{ 58.0}\\
            \bottomrule
          \end{tabular}}
        \caption{  Transferability ($\epsilon \le 16$) comparison of FSA with our attack (SSP). Results are reported for ImageNet-NeurIPS dataset. Lower is better.}
        \label{tab:3}
\end{table}

\section{SSP vs. CDA}
\label{sec:ssp_vs_cda}
We compare our SSP with a recent transferable attack \cite{naseer2019cross}  in Table \ref{tab:4} on MS-COCO validation set using Mask-RCNN. mAP is reported with IoU = 0.5.

\begin{table}[!t]
\centering
\scalebox{0.85}{
        \begin{tabular}{@{}lccc@{}}
            \toprule
            Attack & $\epsilon \le 8/255$ & $\epsilon \le 16/255$ \\
            \midrule
            CDA-ImageNet  & 35.2 & \textbf{8.1}\\
            CDA-Comics & 40.5 & 16.8\\
            CDA-Paintings & 41.7 & 14.8\\
            SSP & \textbf{31.8} & 9.7\\
            \bottomrule
          \end{tabular}}
        \caption{  SSP is compared with CDA \cite{naseer2019cross}. Lower is better.}
        \label{tab:4}
\end{table}

% Segmentation figure
\begin{figure*}[!htp]
\centering
  \begin{minipage}{0.24\textwidth}
  	\centering
    \includegraphics[width=\linewidth, height=3cm]{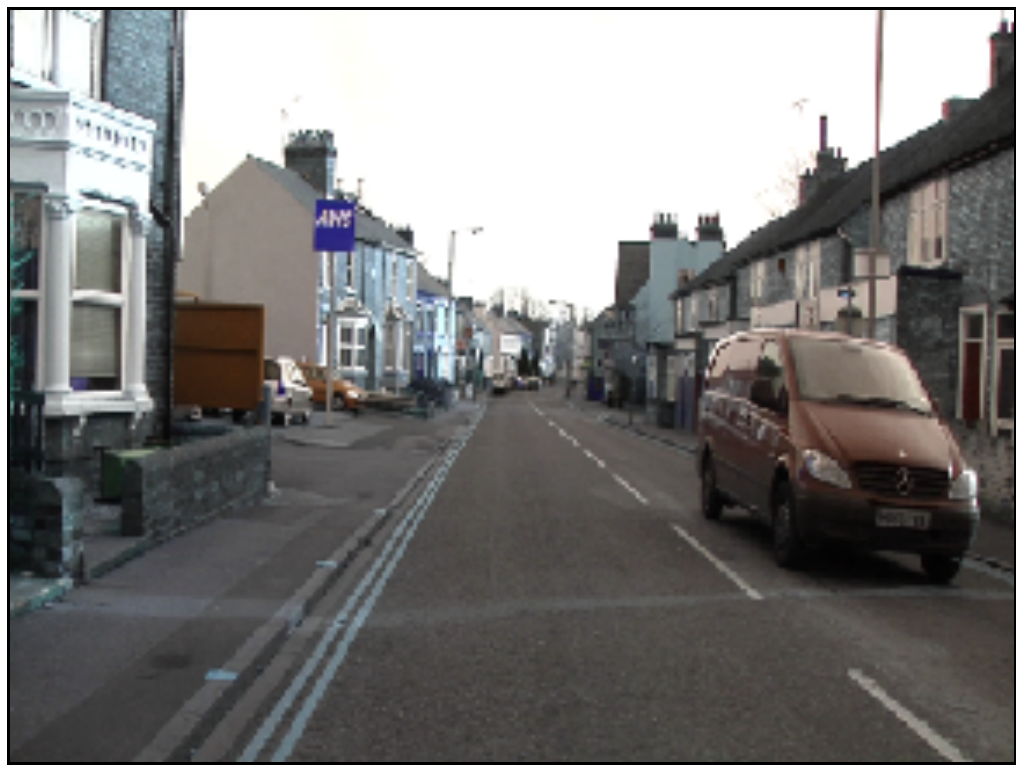}\
     (a) \small Original
  \end{minipage}
  \begin{minipage}{0.24\textwidth}
  	\centering
    \includegraphics[width=\linewidth, height=3cm]{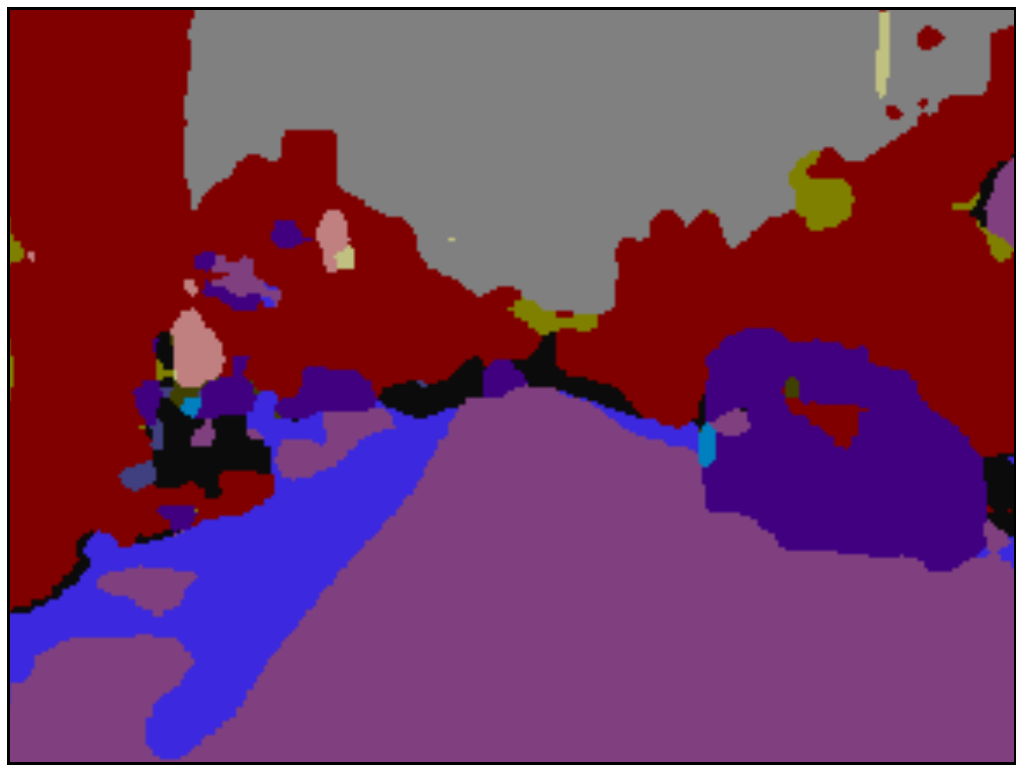}\
      (b) \small Prediction for Original
  \end{minipage}
  \begin{minipage}{0.24\textwidth}
  	\centering
    \includegraphics[width=\linewidth, height=3cm]{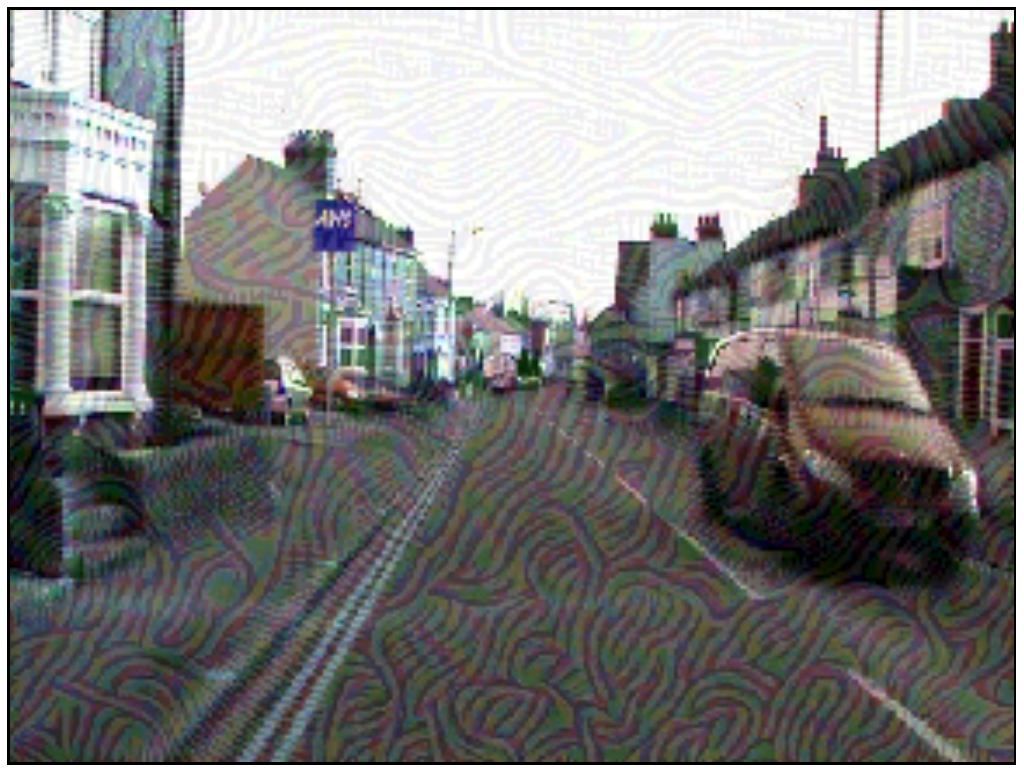}\
     (c) \small Adversarial 
  \end{minipage}
  \begin{minipage}{0.24\textwidth}
  	\centering
    \includegraphics[width=\linewidth, height=3cm]{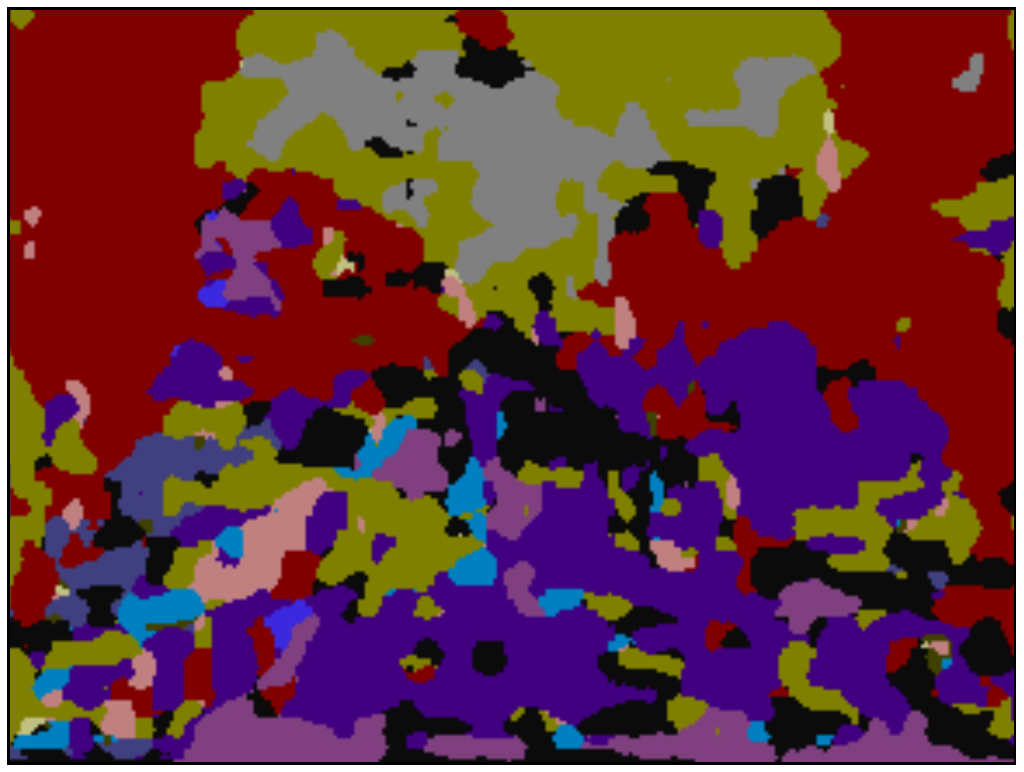}\
      (d) \small Prediction for Adversarial
  \end{minipage}
  %\vspace{-0.2cm}
  \caption{Segnet-Basic output is shown for different images. (a) is the original image, while (b) shows predictions for the original image. (c) is the adversary found by SSP attack, while (d) shows predictions for the adversarial image. Perturbation budget is $l_\infty \le 16$.}
\label{fig:seg_images}
\end{figure*}
%##############################################
% Object detection figure
\begin{figure*}[!htp]
\centering
  \begin{minipage}{0.24\textwidth}
  	\centering
    \vspace{1em}
    \includegraphics[width=\linewidth, height=3cm]{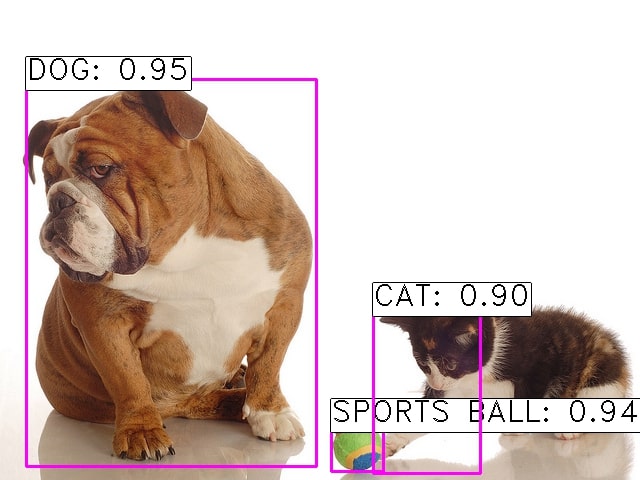}\
     (a) \small  Original
     \vspace{1em}
  \end{minipage}
  \begin{minipage}{0.24\textwidth}
  	\centering
    $l_\infty \le 8$
    \includegraphics[width=\linewidth, height=3cm]{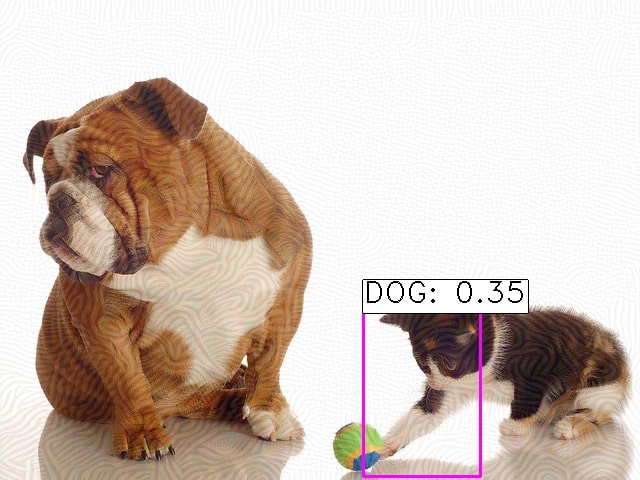}\
      (b) \small  Adversarial
      \vspace{1em}
  \end{minipage}
  \begin{minipage}{0.24\textwidth}
  	\centering
    \vspace{1em}
    \includegraphics[width=\linewidth, height=3cm]{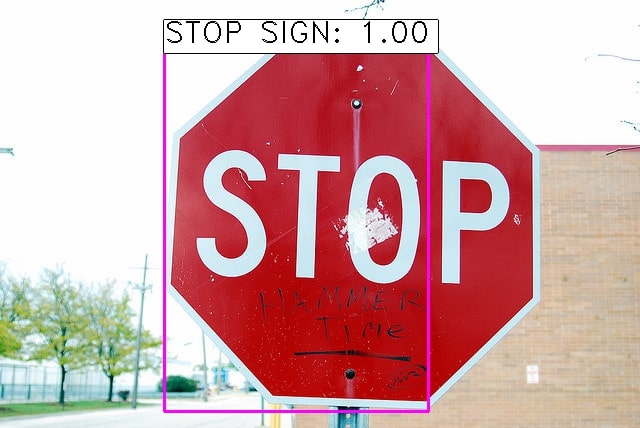}\
     (c) \small  Original
     \vspace{1em}
  \end{minipage}
  \begin{minipage}{0.24\textwidth}
  	\centering
    $l_\infty \le 16$
    \includegraphics[width=\linewidth, height=3cm]{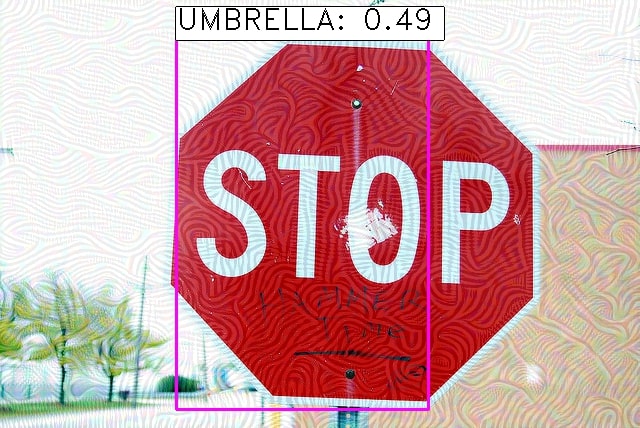}\
      (d) \small  Adversarial
      \vspace{1em}
  \end{minipage}
  %\vspace{-0.4cm}
  \caption{RetinaNet detection results are shown for different images. (a) and (c) show detection for the original images, while (b) and (d) show detection for adversaries found using SSP attack.}
\label{fig:detect_images}
\end{figure*}
%#######################################################

\section{Effect of Input Transformations on SSP Attack}\label{subsec:input-transformation-SSP}
Different input transformations have been proposed to mitigate the adversarial effect. We have tested strength of SSP attack against well studied transformations including:
\begin{itemize}[noitemsep,leftmargin=*,topsep=0em]
\item \emph{JPEG}: This transformation reduces adversarial effect by removing high frequency components in the input image.
\item \emph{Total Variation Minimization (TVM)}: TVM measures small variations thus it can be effective against relatively smaller adversarial perturbations.
\item \emph{Median Filtering (MF)}: This transformation filters out the input image by replacing each pixel with the median of its neighboring pixels.
\end{itemize}
We report our experimental results on segmentation and object detection tasks. 

\textbf{Segmentation:}
 SSP attack created on CAMVID \cite{brostow2009semantic} was able to bring down per pixel accuracy of Segnet-Basic by 47.11\% within $l_\infty \le 16$ (see Table \ref{tab:SSP_segnet} and Figure \ref{fig:seg_images}). JPEG and TVM transformations are slightly effective but only at the cost of drop in  accuracy on benign examples.

\textbf{Object Detection:} RetinaNet \cite{lin2018focal} collapses in the presence of adversaries found by SSP on MS-COCO validation set. Its mean average precision (mAP) with 0.5 intersection over union (IOU) drops from 53.78\% to 5.16\% under perturbation budget $l_\infty \le 16$ (see Table \ref{tab:SSP_ratinanet} and Figure \ref{fig:detect_images}). TVM is relatively more effective compared to other transforms against the SSP.

% IamgeNet-NIPS benign accuracies
\begin{table*}[t]
\small
\centering
\caption{Model accuracies are reported on original data set ImageNet-NIPS containing benign examples only. T-1: top-1 and T-5: top-5 accuracies. Best performances are shown in bold.}
\label{tab:acc-nips}
\vspace{-2ex}
\begin{tabular}{c|p{9ex}<{\centering}|p{9ex}<{\centering}|p{9ex}<{\centering}|p{11ex}<{\centering}|p{9ex}<{\centering}|p{9ex}<{\centering}|p{11ex}<{\centering}|c}
\toprule[0.4mm]
Accuracy &\multicolumn{5}{c|}{Naturally Trained}&\multicolumn{3}{c}{Adv. Trained}\\
\cline{2-9} 
 & Inc-v3 & Inc-v4 & Res-152 & IncRes-v2 & VGG-19 & Adv-v3 & Inc-v3\textsubscript{ens3} & IncRes-v2\textsubscript{ens} \\
\midrule[0.4mm]
\hline
T-1&95.3&97.7&96.1&\textbf{100.0}&85.5&95.1&93.9&97.8\\
\hline
T-5&99.8&99.8&99.9&\textbf{100.0}&96.7&99.4&98.1&99.8\\
\bottomrule[0.4mm]
\end{tabular}
\end{table*}

\begin{table*}[!htp]
\begin{minipage}{0.48\textwidth}
\caption{Segnet-Basic accuracies on CAMVID test set with and without input transformations against SSP. Best performances are shown in bold.}
\centering\setlength{\tabcolsep}{5pt}
\scalebox{0.95}{
\begin{tabular}{|l|l|l|c|}
\hline 
Method & No Attack &\multicolumn{2}{c|}{SSP}\\
\cline{3-4} 
&  & $l_\infty \le 8$ & $l_\infty \le 16$   \\
\hline\hline
No Defense & \textbf{79.70} & 52.48 & 32.59  \\
\hline
JPEG (quality=75) & 77.25&51.76 & 32.44\\
%\hline
JPEG (quality=50) & 75.27& 52.45& 33.16\\
%\hline
JPEG (quality=20) & 68.82 & 53.08& 35.54 \\
\hline
TVM (weights=30) & 73.70 & 55.54 & 34.21 \\ 
%\hline
TVM (weights=10) & 70.38 &\textbf{59.52} & \textbf{34.57}\\ 
\hline
MF (window=3) & 75.65 & 49.18 & 30.52\\ 
\hline
\end{tabular}}

\label{tab:SSP_segnet}
\end{minipage}
\hfill
\begin{minipage}{0.48\textwidth}
\caption{mAP (with IoU = 0.5) of RetinaNet is reported on MS-COCO validation set with and without input transformations against SSP. Best performances are shown in bold.}
\centering\setlength{\tabcolsep}{5pt}
\scalebox{0.95}{
\begin{tabular}{|l|l|l|c|}
\hline 
Method & No Attack &\multicolumn{2}{c|}{SSP}\\
\cline{3-4} 
&  & $l_\infty \le 8$ & $l_\infty \le 16$   \\
\hline\hline
No Defense &\textbf{ 53.78}  & 22.75 & 5.16  \\
\hline
\hline
JPEG (quality=75) & 49.57 & 20.73 & 4.7\\
%\hline
JPEG (quality=50) & 46.36 & 19.89 & 4.33 \\
%\hline
JPEG (quality=20) & 40.04 & 19.13 & 4.58\\
\hline
TVM (weights=30) & 47.06 & 27.63 & 6.36 \\ 
%\hline
TVM (weights=10) & 42.79 & \textbf{32.2}1 & \textbf{9.56}\\ 
\hline
MF (window=3) & 43.48 & 19.59 & 5.05\\ 
\hline
\end{tabular}}

\label{tab:SSP_ratinanet}
\end{minipage}
\end{table*}

\section{Attack Parameters}\label{subsec:attack-parameters}
\begin{figure}[ht]
  \centering
   \vspace{-2.4ex}
    \includegraphics[ width=0.9\linewidth] {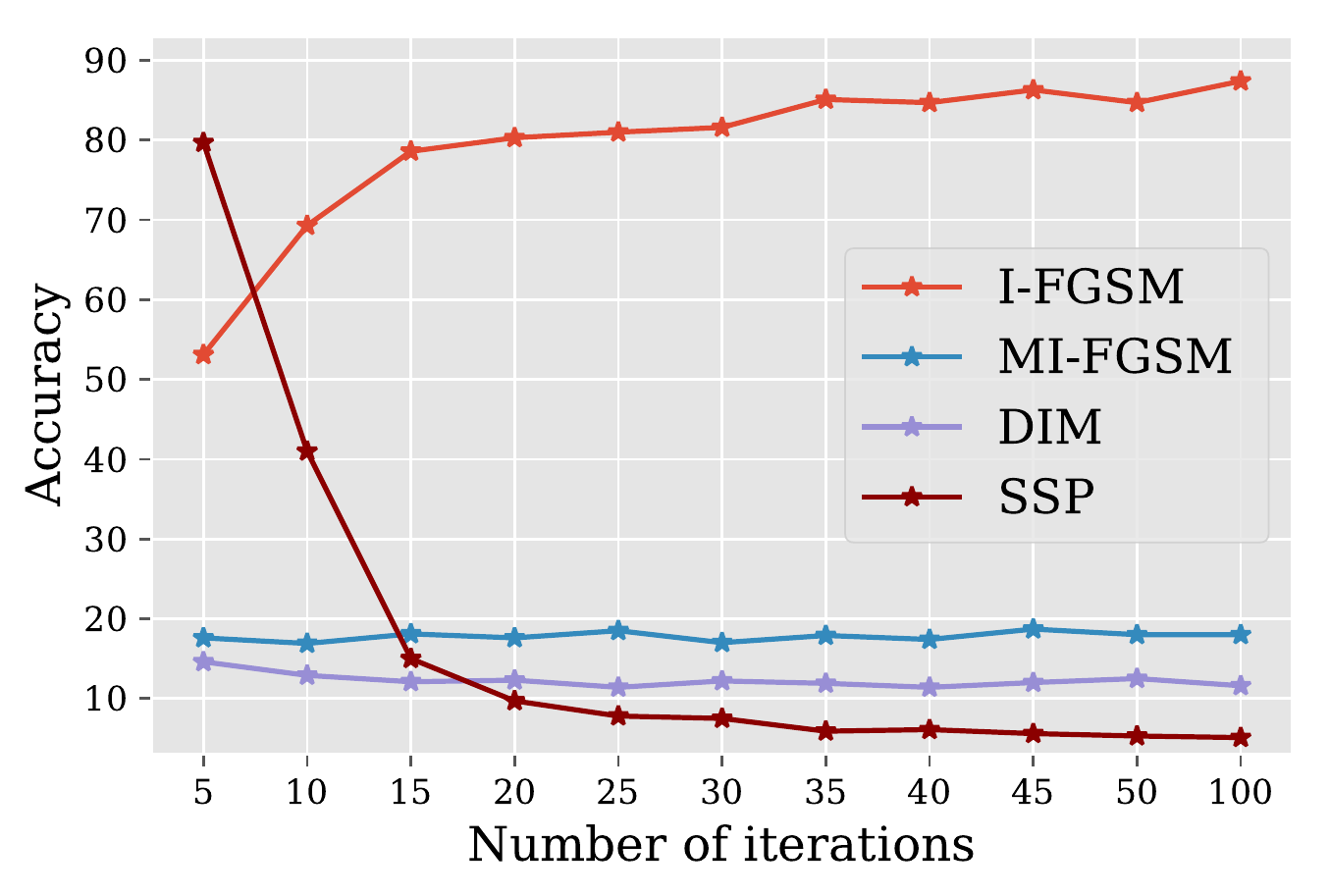}
    \caption{ Accuracy of Inc-v3 for adversaries created on VGG-16 by different attacks. SSP's strength increases with number of iterations, in contrast to MI-FGSM and DIM.}
    \label{fig:SSP_iterations_analysis}
    %\vspace{2ex}
\end{figure}

For FGSM, we use a step size of $16$. For R-FGSM, we take a step of size $\alpha{=}16/3$ in a random direction and then a gradient step of size $16{-}\alpha$ to maximize model loss. The attack methods, I-FGSM, MI-FGSM and DIM, are run for $10$ iterations. The step size for these attacks is set to $1.6$, as per the standard practice. The momentum decay factor for MI-FGSM is set to 1. This means that attack accumulates all the previous gradient information to perform the current update and is shown to have the best success rate \cite{Dong_2018_CVPR}. For DIM, the transformation probability is set to 0.7. In the case of FFF \cite{mopuri-bmvc-2017}, we train the adversarial noise for 10K iterations to maximize the response at the activation layers of VGG-16 \cite{Simonyan14verydeep}. For the SSP,  we used VGG-16 \cite{Simonyan14verydeep} conv3-3 feature map as the feature loss. Since SSP generation approach maximizes loss w.r.t a benign example, it does not suffer from the over-fitting problem. We run SSP approach for the maximum number of 100 iterations. The transferability of different attacks is compared against the number of iterations in Figure~\ref{fig:SSP_iterations_analysis}. MI-FGSM and DIM quickly reach to their full potential within ten iterations. The strength of I-FGSM strength decreases, while feature distortion strength (SSP) increases with the number of attack iterations. Top-1 (T-1) and Top-5 (T-5) accuracies of Imagenet trained models on NeurIPS dataset are reported in Table \ref{tab:acc-nips}.

%#################################################
% attacks vs iterations
%##############################################

\section{Generalization to Unseen Attacks}\label{subsec:NRP-genralization-demo}
\begin{figure*}[!h]
\centering
% \vspace{40ex}
\includegraphics[width=\linewidth, keepaspectratio]{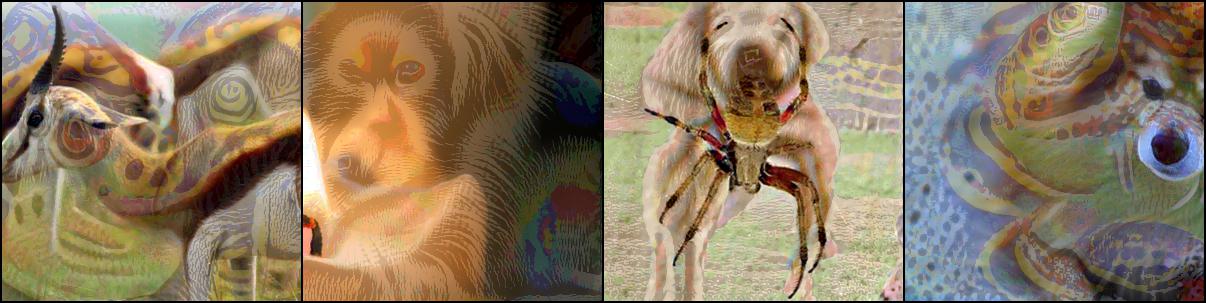}
Adversaries produced by SSP using adversarilly robust features \cite{engstrom2019adversarial}.
\includegraphics[width=\linewidth, keepaspectratio]{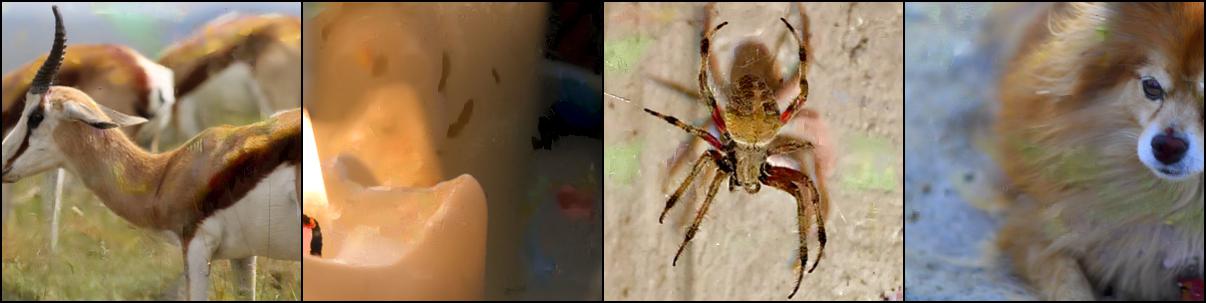}
 Purified adversaries by NRP.
\caption{NRP is capable to remove these difficult adversaries where adversarial image is in-painited into the clean image. Untargetted adversaries are created by applying SSP to feature space of adversarially trained ResNet50 \cite{engstrom2019adversarial}. Perturbation budget is set to $l_\infty \le 16$.}
 \label{fig:supplementary-demo-nrp-robust}
\end{figure*}

\begin{figure*}[ht]
\centering
\includegraphics[width=\linewidth, keepaspectratio]{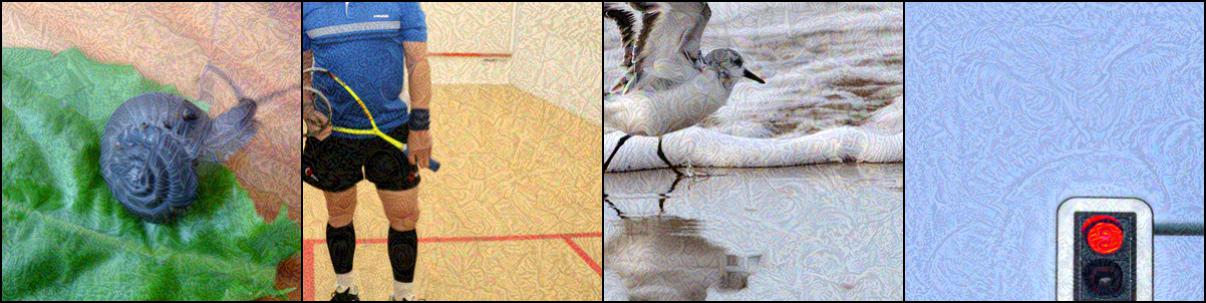}
Adversaries produced by DIM \cite{xie2018improving}
\includegraphics[width=\linewidth, keepaspectratio]{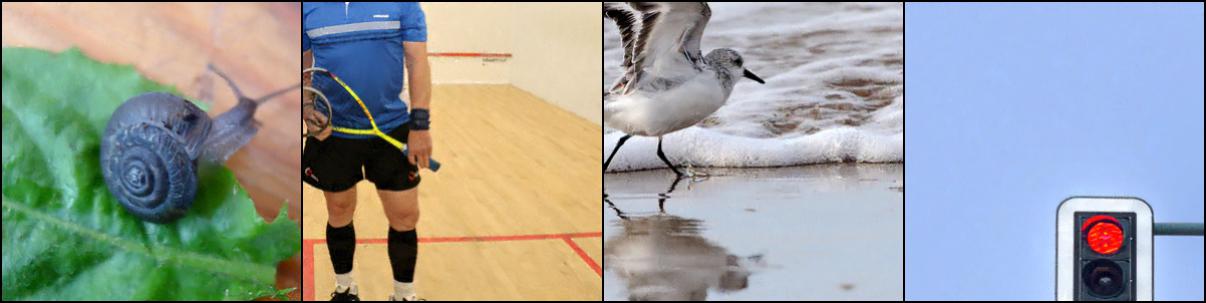}
 Purified adversaries by NRP
\includegraphics[width=\linewidth, keepaspectratio]{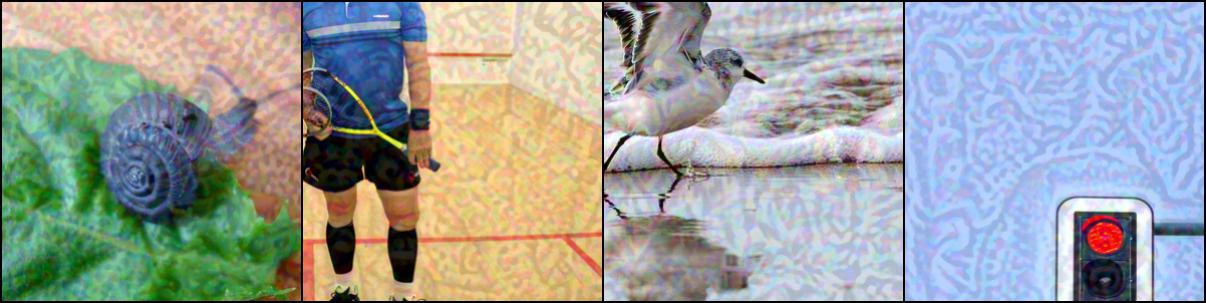}
Adversaries produces by DIM$_{TI}$ \cite{dong2019evading}
\includegraphics[width=\linewidth, keepaspectratio]{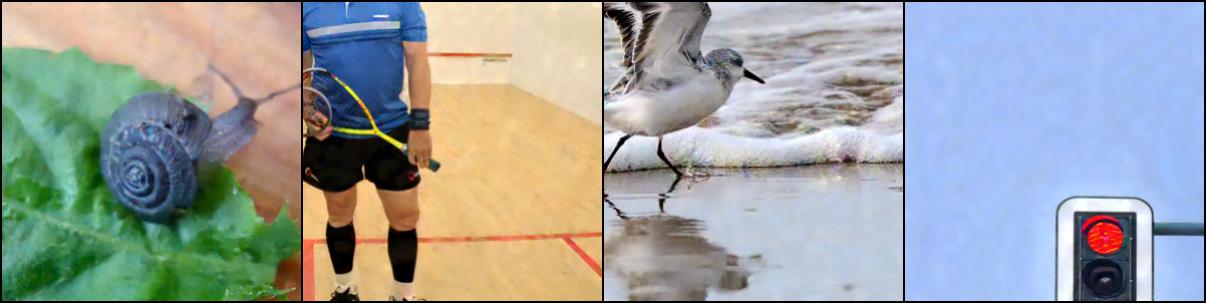}
Purified adversaries by NRP
 
\caption{NRP removes diverse patterns produces by DIM \cite{xie2018improving} and translation-invariant attacks \cite{dong2019evading} to a great extent. Untargetted adversaries are created by ensemble of \textbf{ensemble} of Inc-v3, Inc-v4, IncRes-v2, and Res-152. Perturbation budget is set to $l_\infty \le 16$.}
 \label{fig:supplementary-demo-nrp-ens}
\end{figure*}

\begin{figure*}[ht]
\centering
\includegraphics[width=\linewidth, keepaspectratio]{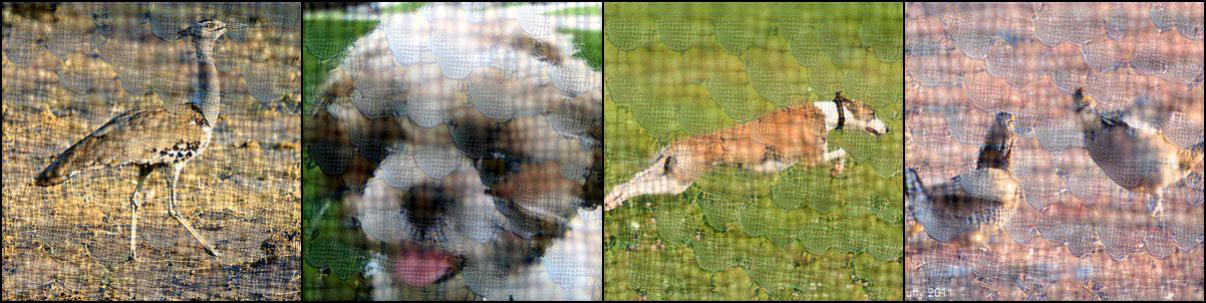}
Adversaries produced by CDA \cite{naseer2019cross} - ImageNet
\includegraphics[width=\linewidth, keepaspectratio]{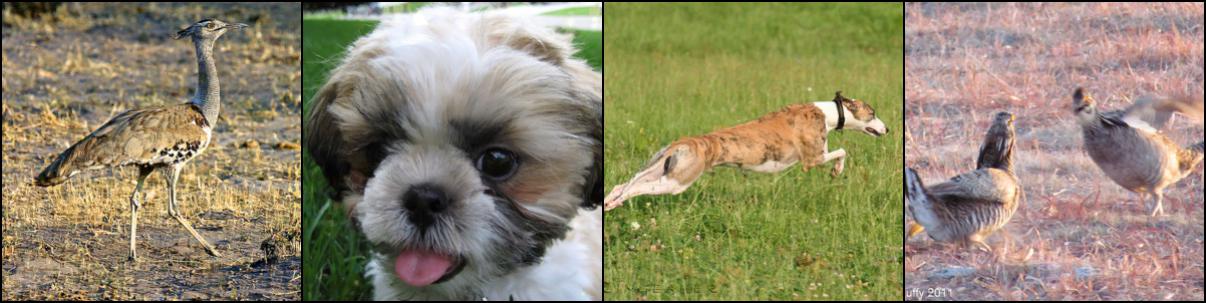}
Purified adversaries by NRP
\includegraphics[width=\linewidth, keepaspectratio]{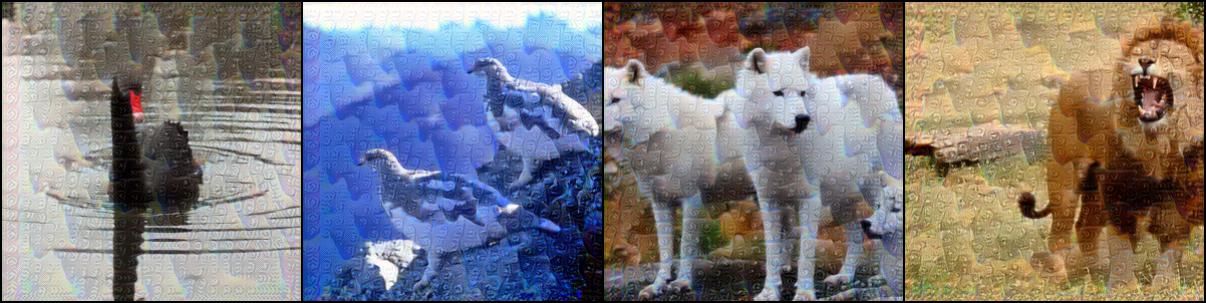}
Adversaries produced by CDA \cite{naseer2019cross} - Paintings
\includegraphics[width=\linewidth, keepaspectratio]{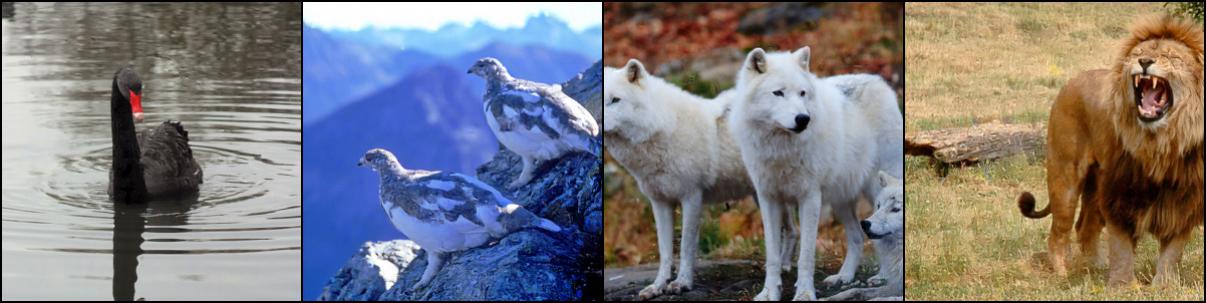}
Purified adversaries by NRP
 
\caption{Our defense successfully able to recover original samples from unseen adversarial patterns. These are untargeted adversaries produced by CDA \cite{naseer2019cross} trained against Inc-v3 on ImageNet and Paintings. Perturbation budget is set to $l_\infty \le 16$.}
 \label{fig:supplementary-demo-nrp-incv3}
\end{figure*}

\begin{figure*}[ht]
\centering
  \begin{minipage}{.48\textwidth}
  	\centering
  	 Adversarial
  	 % lelf lower right up 
    \includegraphics[ width=\linewidth, height=5cm]{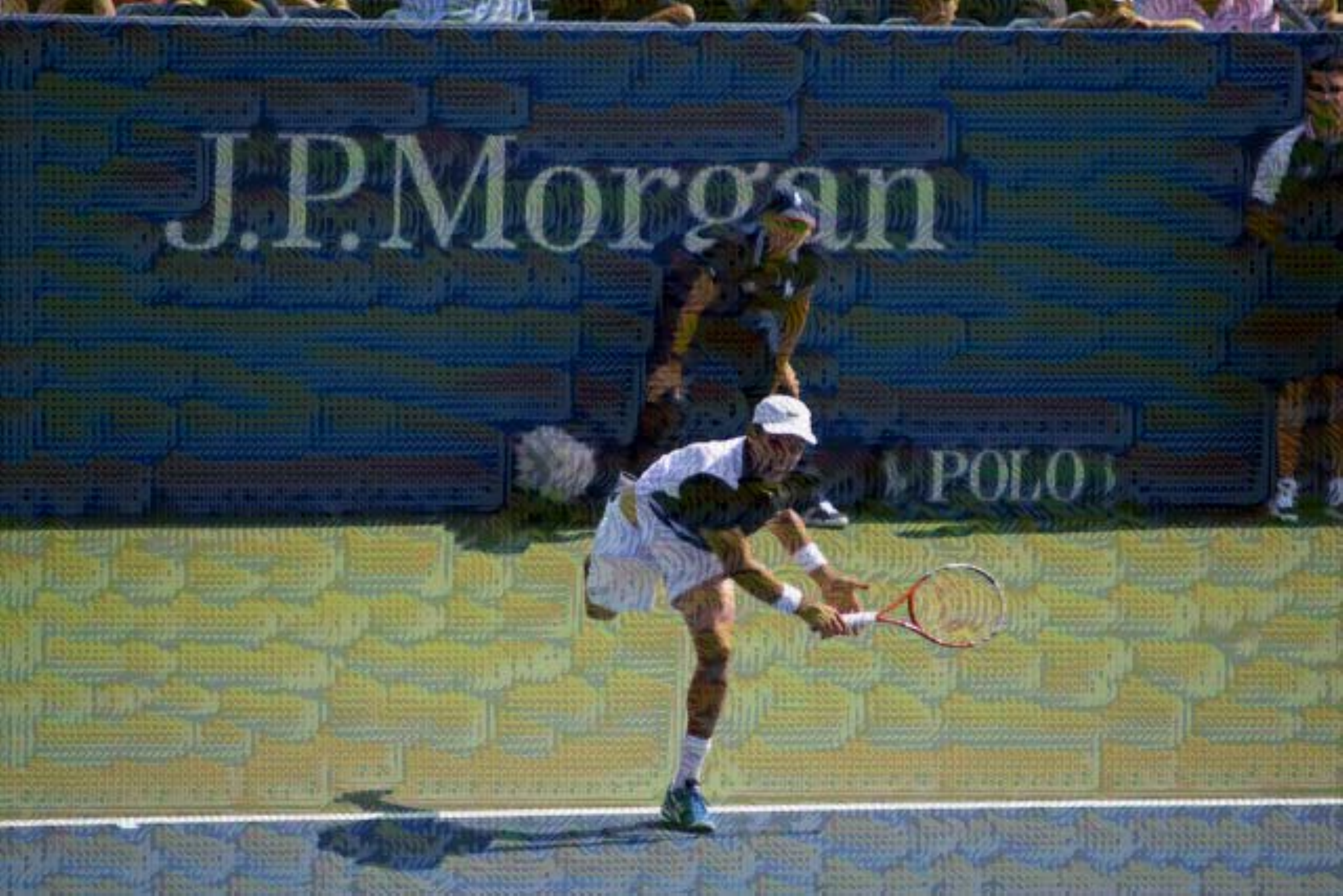}\
  \end{minipage}
    \begin{minipage}{.48\textwidth}
  	\centering
  	Prediction for Adversarial
  	 % lelf lower right up 
    \includegraphics[trim= 0mm 0mm -10mm 0mm, width=\linewidth, height=5cm]{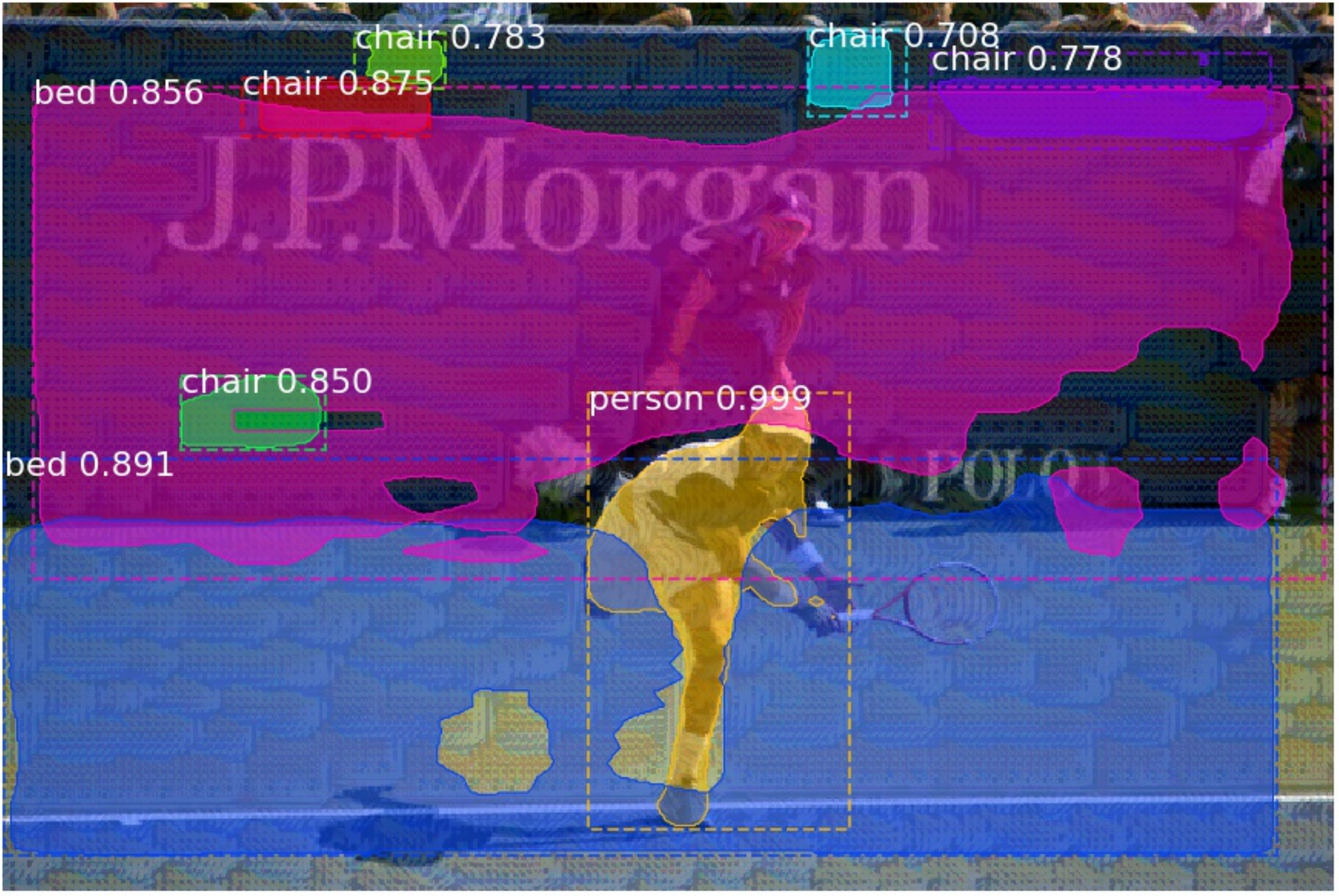}\
  \end{minipage} \\
  \begin{minipage}{.48\textwidth}
  	\centering
  	Purified
  	 % lelf lower right up 
    \includegraphics[ width=\linewidth, height=5cm]{images/supplimentary/CDA_demo/rcnn/fixed/000000000885.pdf}\
     
  \end{minipage}
    \begin{minipage}{.48\textwidth}
  	\centering
  	 Prediction for Purified
  	 % lelf lower right up 
    \includegraphics[trim= 0mm 0mm 14mm 0mm, clip, width=\linewidth, height=5cm]{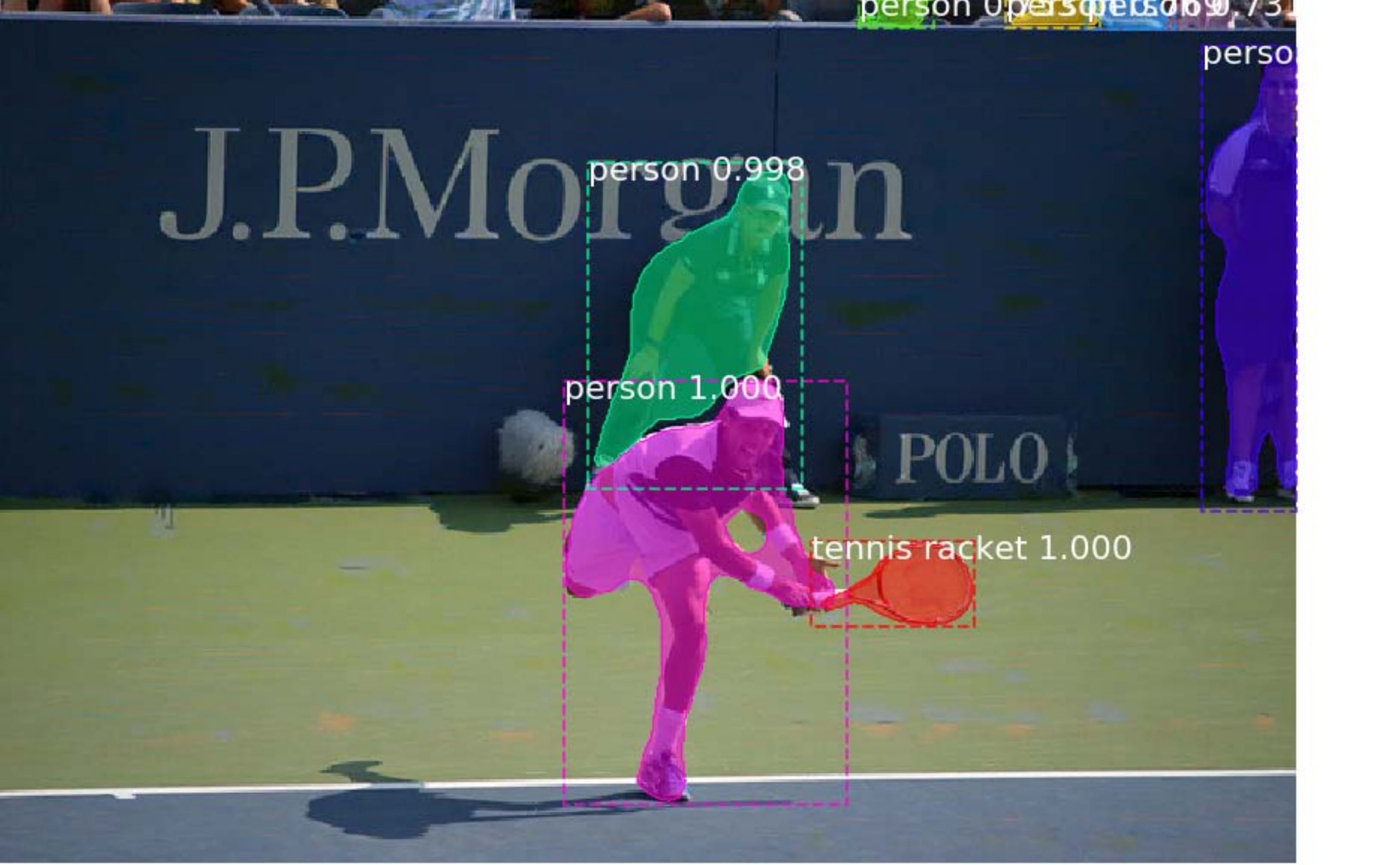}\
   
  \end{minipage}\\

    \begin{minipage}{.48\textwidth}
  	\centering
  	Adversarial
  	 % lelf lower right up 
    \includegraphics[ width=\linewidth, height=5cm]{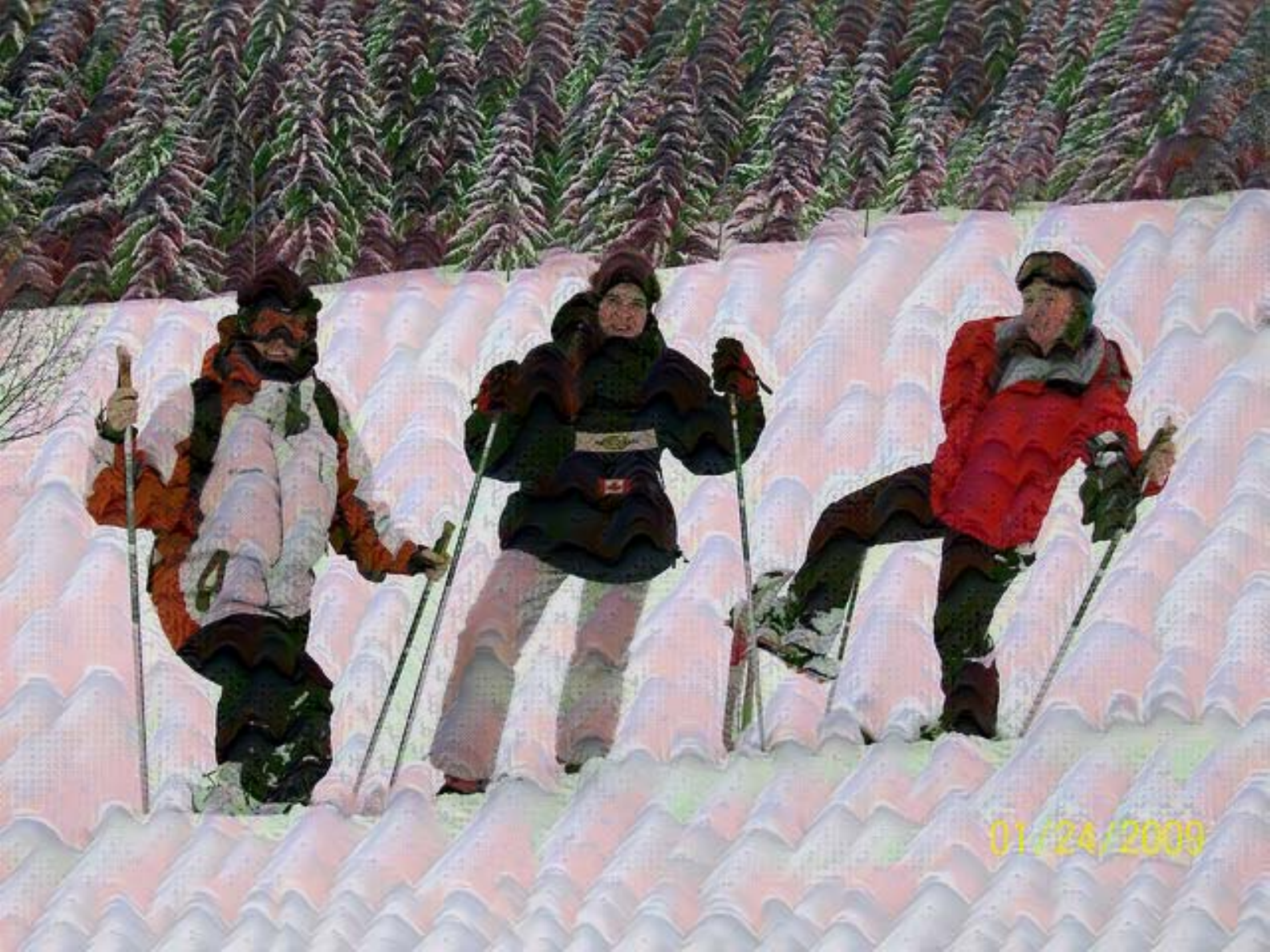}\
  \end{minipage}
    \begin{minipage}{.48\textwidth}
  	\centering
  	Prediction for Adversarial
  	 % lelf lower right up 
    \includegraphics[trim= 0mm 0mm -10mm 0mm, width=\linewidth, height=5cm]{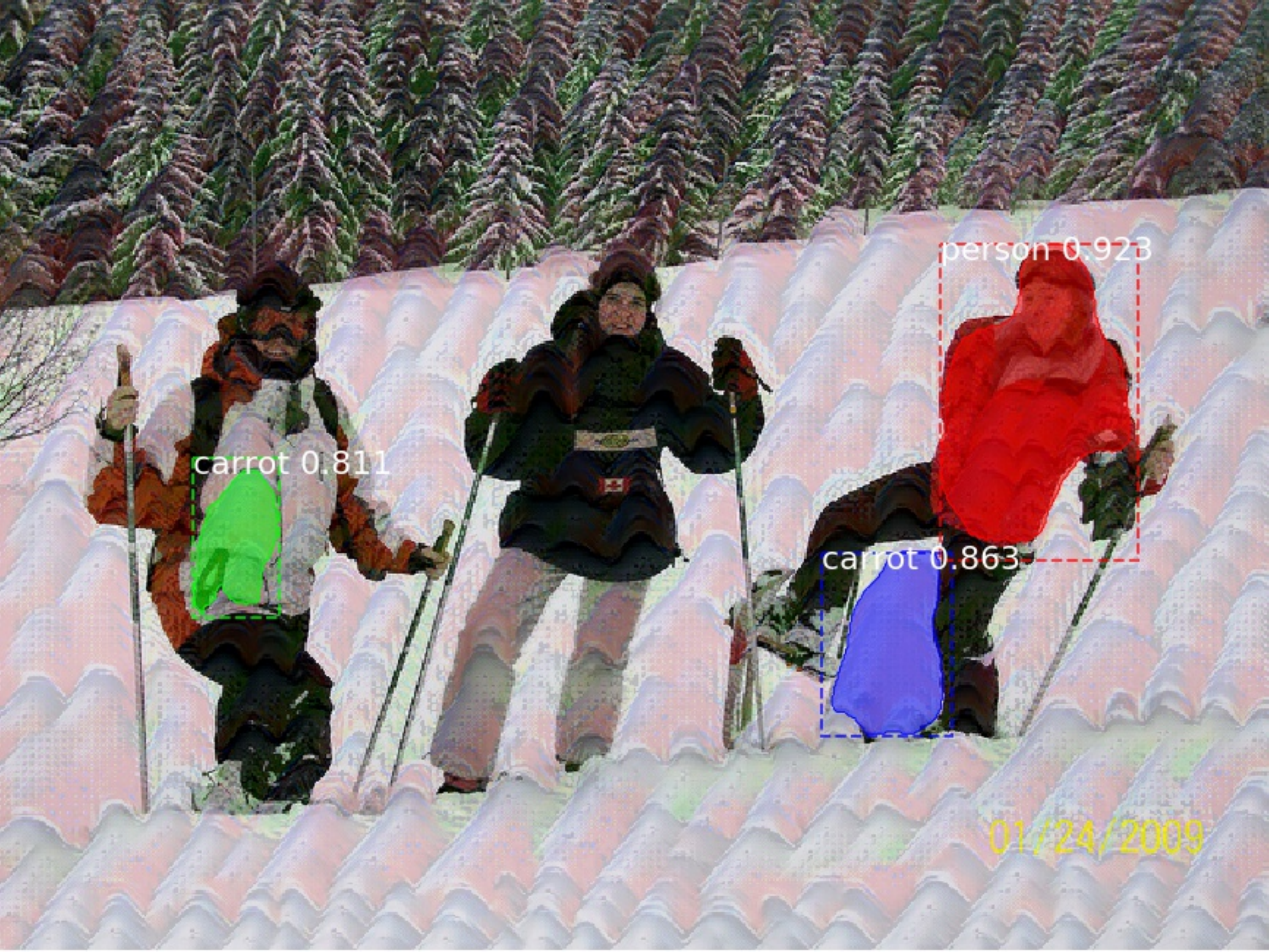}\
  \end{minipage} \\
  \begin{minipage}{.48\textwidth}
  	\centering
  	Purified
  	 % lelf lower right up 
    \includegraphics[ width=\linewidth, height=5cm]{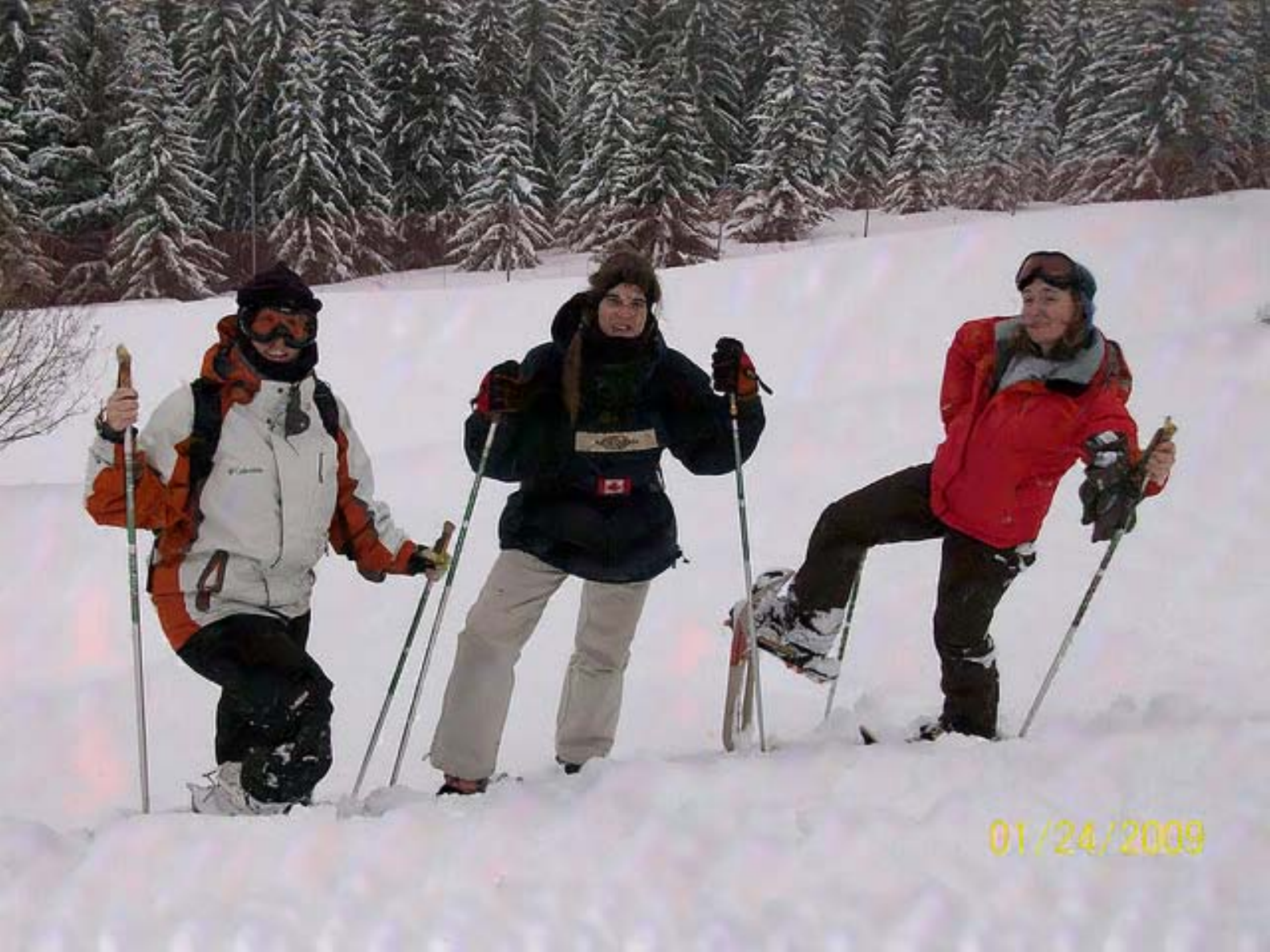}\
  \end{minipage}
    \begin{minipage}{.48\textwidth}
  	\centering
  	Prediction for Purified
  	 % lelf lower right up 
    \includegraphics[trim= 0mm 0mm -10mm 0mm, clip, width=\linewidth, height=5cm]{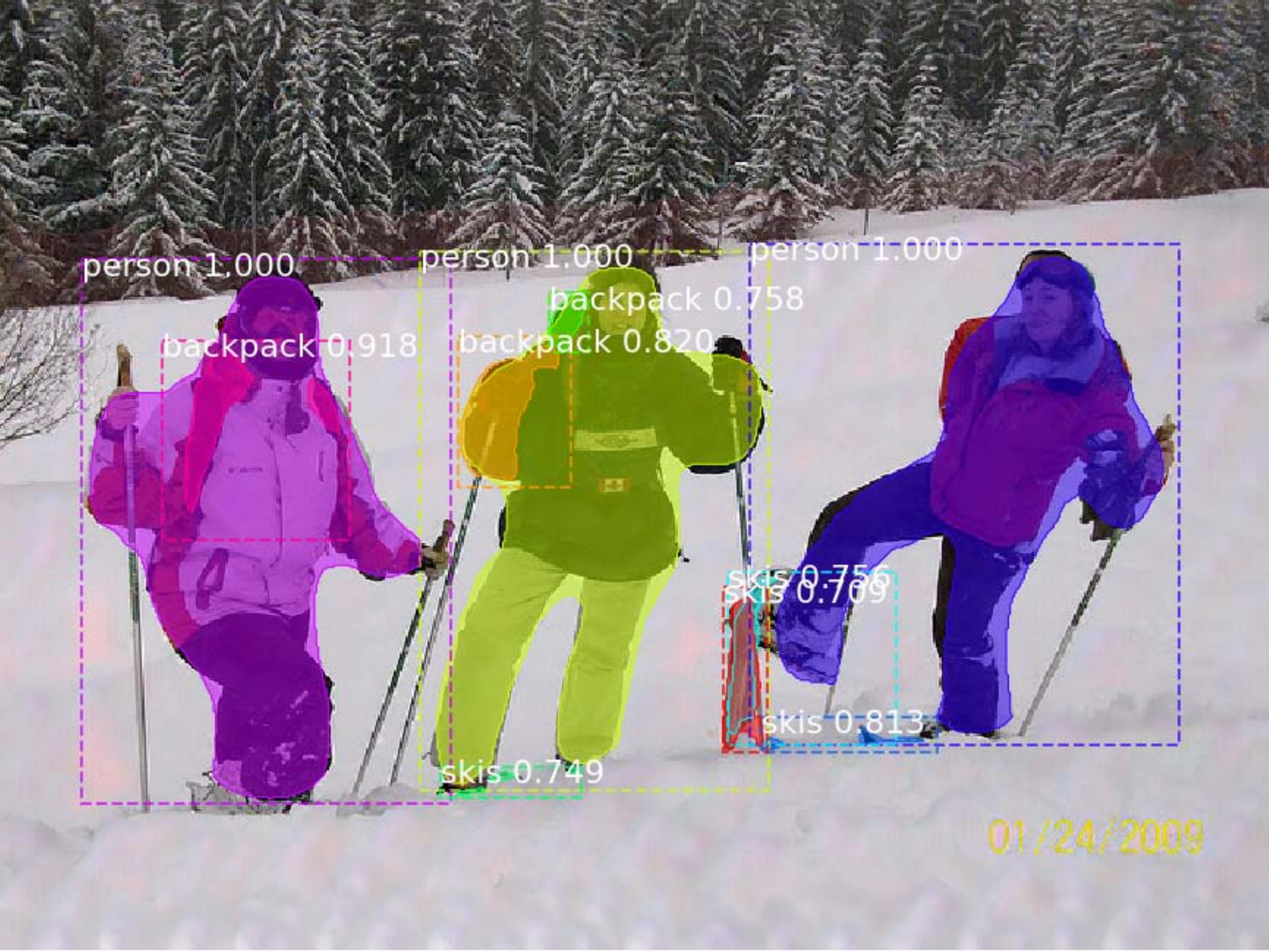}\
  \end{minipage}
 
\caption{ Adversaries generated by CDA \cite{naseer2019cross} reduce Mask-RCNN \cite{He2017MaskR} performance. NRP successfully removes adversarial perturbations and greatly stabilizes Mask-RCNN predictions.}
 \label{fig:supplementary-demo-nrp-rcnn}
\end{figure*}

We show visual demonstration (see Figures \ref{fig:supplementary-demo-nrp-robust}, \ref{fig:supplementary-demo-nrp-ens}, \ref{fig:supplementary-demo-nrp-incv3} and \ref{fig:supplementary-demo-nrp-rcnn}) of how our defense, NRP, trained using SSP attack is able to generalize on the variety of unseen perturbations created by different attack algorithms. NRP successfully removes the perturbations that it never saw during training.

\begin{itemize}[leftmargin=*]
    \item Figure \ref{fig:supplementary-demo-nrp-robust} shows adversaries coming from adversarially robust model. It's the most difficult case as perturbations does not resemble to a noisy patter rather represent meaningful structured pattern that are in-painted into the clean image. NRP's ability to remove such difficult patterns shows that our defense can separate  the original signal from the adversarial one. 
    \item NRP has no difficulty in removing thick patterns introduced by DIM  or smooth perturbations of DIM-TI attacks (Figure \ref{fig:supplementary-demo-nrp-ens}).
    %\item NRP can easily separate original signal from cross-domain adversarial pattern (Figures \ref{fig:supplementary-demo-nrp-incv3} and \ref{fig:supplementary-demo-nrp-rcnn}).
\end{itemize}

\vspace{-3cm}

\end{document}

%% file: tab_def_1.tex
\begin{table}[t]
\setlength{\tabcolsep}{6.3pt}

     \caption{\small Robustness of different defense methods against state-of-the-art black-box attacks (\emph{lower is better}). IncRes-v2\textsubscript{ens} is used as backbone model following \cite{dong2019evading}. NRP significantly reduces the attack success rate. Adversaries ($\epsilon \le 16$) are created against Inc-v3, Inc-v4, IncRes-v2, Res-v2-152 and Ensemble.}
\vspace{-.4em}

\label{tab:NRP-main-defense-results}
\scalebox{0.85}{
    \begin{tabular}{L{0.4cm}|l|cc|cc|cc}
    \toprule[0.4mm]
    & \multirow{2}{*}{\textbf{Defenses}} & \multicolumn{6}{c}{\textbf{Attacks}}\\
     \cline{3-8} 
        &  & \rotatebox{90}{\small FGSM} & \rotatebox{90}{\small FGSM$_{TI}$} & \rotatebox{90}{ \small MIFGSM} & \rotatebox{90}{\small MIFGSM$_{TI}$} & \rotatebox{90}{\small DIM} &  \rotatebox{90}{\small DIM$_{TI}$}  \\
        \midrule[0.4mm]
       
      \parbox[t]{2mm}{\multirow{6}{*}{\rotatebox[origin=c]{90}{Inc-v3}}}   & JPEG %(\venue{ICLR'18}) 
      \cite{guo2018countering} & 19.9&25.5&20.3&28.2&30.7&37.0 \\
      %\cmidrule(lr){2-8}
        & TVM %(\venue{ICLR'18}) 
        \cite{guo2018countering} & 18.8&30.7&19.4&34.9&24.4&44.2\\
        %\cmidrule(lr){2-8}
         & NIPS-r3 %(\venue{NeurIPS'17}) 
         \cite{NIPS-r3} &9.8&24.5&12.9&26.7&18.0&41.4\\
        %\cmidrule(lr){2-8}
         & R\&P %(\venue{ICLR'17}) 
         \cite{xie2017mitigating} &6.5&19.8&8.7&23.9&13.3&36.8 \\
        %\cmidrule(lr){2-8}
        & HGD %(\venue{CVPR'18}) 
        \cite{Liao_2018_CVPR} &\textbf{2.1}&18.4&6.9&25.7&9.7&38.3  \\
        &APE-GAN %(\venue{ICASSP'19}) 
        \cite{shen2017ape} &19.6&28.0&17.9&30.4&23.6&38.6\\
        &SR %(\venue{TIP'19}) 
        \cite{mustafa2019image} &23.0&36.7&23.6&38.3&32.5&49.0\\
        \cline{2-8}
        & NRP & 3.2 &\textbf{4.8}&\textbf{4.5}&\textbf{9.1}&\textbf{5.1}&\textbf{11.0}\\
        \hline
        \parbox[t]{2mm}{\multirow{6}{*}{\rotatebox[origin=c]{90}{Inc-v4}}}   & JPEG %(\venue{ICLR'18}) 
        \cite{guo2018countering} & 21.8&27.9&26.0&31.6&38.6&43.5 \\
        %\cmidrule(lr){2-8}
        & TVM %(\venue{ICLR'18}) 
        \cite{guo2018countering} & 19.9 & 31.8& 24.8 & 38.4 & 29.1 & 45.6\\
        %\cmidrule(lr){2-8}
        & NIPS-r3 %(\venue{NeurIPS'17}) 
        \cite{NIPS-r3} &11.5&24.6&15.6&29.5&14.1&41.9\\
        %\cmidrule(lr){2-8}
        & R\&P %(\venue{ICLR'17}) 
        \cite{xie2017mitigating}  &7.9&21.6&12.1&28.0&17.2&39.3 \\
        %\cmidrule(lr){2-8}
        & HGD %(\venue{CVPR'18}) 
        \cite{Liao_2018_CVPR} & \textbf{2.6} & 18.1 & 9.6 & 27.8 & 32.4 & 58.7 \\
         &APE-GAN %(\venue{ICASSP'19}) 
         \cite{shen2017ape}&21.1&28.8&20.7&32.8&25.0&39.0\\
          &SR %(\venue{TIP'19}) 
          \cite{mustafa2019image}&25.3&34.1&29.2&42.3&39.3&52.3\\
        \cline{2-8}
        & NRP & 3.1&\textbf{4.4}&\textbf{4.8}&\textbf{10.3}&\textbf{5.2}&\textbf{12.5}\\
        \hline
                
        \parbox[t]{2mm}{\multirow{6}{*}{\rotatebox[origin=c]{90}{IncRes-v2}}}   & JPEG %(\venue{ICLR'18}) 
        \cite{guo2018countering} & 24.7 & 32.4&31.6&45.9&47.2&55.7 \\
        %\cmidrule(lr){2-8}
        & TVM %(\venue{ICLR'18}) 
        \cite{guo2018countering} & 23.4&38.5&34.4&55.4&41.7&66.2\\
        %\cmidrule(lr){2-8}
        & NIPS-r3 %(\venue{NeurIPS'17}) 
        \cite{NIPS-r3} &13.3&31.4&22.7&46.2&37.6&61.5\\
        %\cmidrule(lr){2-8}
        & R\&P %(\venue{ICLR'17}) 
        \cite{xie2017mitigating}  & 9.9& 28.1&18.6&45.2&30.2&61.4\\
        %\cmidrule(lr){2-8}
        & HGD %(\venue{CVPR'18}) 
        \cite{Liao_2018_CVPR} & 3.9 & 25.4&19.6&45.1&32.4&58.7 \\
         &APE-GAN %(\venue{ICASSP'19}) 
         \cite{shen2017ape}&24.7&36.8&30.4&50.5&36.3&60.5\\
          &SR %(\venue{TIP'19})
          \cite{mustafa2019image}&27.6&42.4&42.6&62.1&54.3&72.2\\
        \cline{2-8}
        & NRP & \textbf{3.5}&\textbf{6.9}&\textbf{7.6}&\textbf{18.7}&\textbf{7.5}&\textbf{20.8}\\
        \hline
        
        \parbox[t]{2mm}{\multirow{6}{*}{\rotatebox[origin=c]{90}{Res-v2-152}}}   & JPEG %(\venue{ICLR'18}) 
        \cite{guo2018countering} &24.0&32.7&31.2&38.3&42.4&50.8  \\
        %\cmidrule(lr){2-8}
        & TVM %(\venue{ICLR'18}) 
        \cite{guo2018countering} & 22.0&38.1&24.5&41.2&36.8&55.7\\
        %\cmidrule(lr){2-8}
        & NIPS-r3 %(\venue{NeurIPS'17}) 
        \cite{NIPS-r3} &12.5&30.1&18.0&34.4&34.4&52.9\\
        %\cmidrule(lr){2-8}
        & R\&P %(\venue{ICLR'17}) 
        \cite{xie2017mitigating}  &8.6&27.4&14.6&31.1&26.4&50.4 \\
        %\cmidrule(lr){2-8}
        & HGD %(\venue{CVPR'18}) 
        \cite{Liao_2018_CVPR} &3.6&24.4&15.1&31.8&32.6&51.8  \\
         &APE-GAN %(\venue{ICASSP'19}) 
         \cite{shen2017ape}&24.3&37.1&23.2&38.6&34.3&53.8\\
          &SR %(\venue{TIP'19}) 
          \cite{mustafa2019image}&26.3&41.8&30.2&49.2&48.4&63.9\\
        \cline{2-8}
        & NRP & \textbf{3.4}&\textbf{6.5}&\textbf{5.8}& \textbf{11.9}&\textbf{6.3}&\textbf{17.8}\\
        \hline
        
     \parbox[t]{2mm}{\multirow{7}{*}{\rotatebox[origin=c]{90}{Ensemble}}}   
    & JPEG %(\venue{ICLR'18}) 
    \cite{guo2018countering}  & 38.1 & 43.3&67.7&77.2&82.5&83.4  \\
     %\midrule
    & TVM %(\venue{ICLR'18}) 
    \cite{guo2018countering}  & 30.0&39.8&50.1&72.1&64.1&79.8\\
    %\midrule
    & NIPS-r3 %(\venue{NeurIPS'17}) 
    \cite{NIPS-r3} &19.8&33.9&43.9&71.4&63.7&83.1\\
    %\midrule 
    & R\&P %(\venue{ICLR'17}) 
    \cite{xie2017mitigating}  &13.8&31.2&32.8&68.3&51.7&81.4 \\
    %\midrule 
    & HGD %(\venue{CVPR'18}) 
    \cite{Liao_2018_CVPR} &  4.9 & 29.9&38.6&73.3&57.7&82.6\\
    & APE-GAN %(\venue{ICASSP'19}) 
    \cite{shen2017ape} &32.0&42.1&44.6&69.3&59.6&74.5\\
    & SR %(\venue{TIP'19}) 
    \cite{mustafa2019image}&38.1&45.8&65.2&79.9&79.3&84.9\\
    \hline
    & NRP & \textbf{3.7} &\textbf{ 7.9} & \textbf{10.1} &\textbf{ 27.8} & \textbf{11.4} & \textbf{31.9}\\
\bottomrule[0.4mm]
\end{tabular}
    }\vspace{-1em}
\end{table}

%% file: fig_tab_cross_defense.tex
\begin{table}[t]
\setlength{\tabcolsep}{2pt}
\footnotesize
\caption{\small NRP generalizability across different adversarial attacks. Classification model is defended against CDA trained against Inc-v3 while detection and segmentation models are defended against CDA trained against Res-v2-152 (\emph{higher is better}). 
(q=quantity, w=weights, win=window size)} \vspace{-.4em}
    \begin{tabular}{l|c|c|c|c|c|c|c}
    \toprule[0.4mm]
     \multicolumn{8}{c}{\small \textbf{Classification:} Defending IncRes-v2\textsubscript{ens} \cite{tramer2017ensemble} against CDA \cite{naseer2019cross}} \\ \midrule[0.4mm]
    \hline 
    Method & No  &\multicolumn{2}{c|}{ImageNet}&\multicolumn{2}{c|}{Comics}&\multicolumn{2}{c}{Paintings}\\
    \cline{3-8} 
    & Attack & $\footnotesize l_\infty {\le} 8$ & $\footnotesize l_\infty {\le} 16$ & $\footnotesize l_\infty {\le} 8$ & $\footnotesize l_\infty {\le} 16$& $\footnotesize l_\infty {\le} 8$ & $\footnotesize l_\infty {\le} 16$  \\
    \hline
    No Defense &\textbf{ 97.8} & 83.0 & 30.9 &94.0& 56.6&71.6& 23.7 \\
    \hline
    JPEG (q=75) & 97.6& 74.9 & 18.6&90.1& 42.6&68.0& 18.0\\
    JPEG (q=50) & 96.2& 74.2 &19.0 &90.1&43.4 &66.0& 19.2 \\
    JPEG (q=20) & 94.1&73.4&  21.7&87.0&51.3 &62.7& 18.8\\
    TVM (w=10) & 93.1 &82.3& 30.2&91.0&77.2 &72.7& 27.4 \\
    TVM (w=30) & 96.0&81.1&  27.3&93.4& 66.4&70.6& 24.1\\
    MF (win=3) & 95.4 &77.3& 27.7 &92.4& 66.8&65.0& 22.1 \\
    \hline
    NRP & 95.6 & \textbf{95.7}&\textbf{96.0}&\textbf{95.4} &\textbf{94.2}&\textbf{ 95.3} & \textbf{94.1}\\ \hline \midrule[0.4mm]

    \multicolumn{8}{c}{\small\textbf{Detection:} Defending Mask-RCNN \cite{He2017MaskR} against CDA~\cite{naseer2019cross}} \\ \midrule[0.4mm]
    No Defense & \textbf{59.9} & 35.2&8.1&40.5&16.8&41.7&14.8 \\
    \hline
    JPEG (q=75) & 57.6 &41.3&11.9 &41.6& 19.4&44.5&18.3 \\
    JPEG (q=50) & 54.6 &41.7&14.5 &39.5& 18.5&47.7&19.9  \\
    JPEG (q=20) & 39.7 &30.7& 15.1&28.2&14.7 &30.5&15.3 \\
    TVM (w=10) & 54.1 &32.1&14.3 &40.5&28.9 &37.6&21.5  \\
    TVM (w=30) & 58.0 &39.9&10.1 &46.8&21.0 &45.4&17.2\\
    MF (win=3) & 54.7 &32.1&9.0 &41.1&20.4 &37.6&15.2 \\
    \hline
    NRP & 54.4 & \textbf{51.5}&\textbf{50.3}&\textbf{53.5}&\textbf{53.7}&\textbf{53.2}&\textbf{54.3}\\
    \hline
    \midrule[0.4mm]
    \multicolumn{8}{c}{\small\textbf{Segmentation:} Mask-RCNN \cite{He2017MaskR} defense against CDA \cite{naseer2019cross}} \\ \midrule[0.4mm] \hline
    
    No Defense & \textbf{56.8} & 32.4&7.3&37.6&15.5&39.1&13.8   \\
     \hline
    JPEG (q=75) &54.4 &38.5&11 &38.5&17.8 &41.7&16.9 \\
    JPEG (q=50) & 51.5&38.9&13.4 &36.6&17.3 &40&18.2  \\
    JPEG (q=20) & 37.1&28.8&14.0 &26.3& 13.8&28.3& 14.3\\
    TVM (w=10) &50.8 &29.8& 13.2&37.6& 26.6&34.9&19.8  \\
    TVM (w=30) & 54.4&37.1&9.3 &43.7&19.3 &42.3&15.9\\
    MF (win=3) & 51.5&29.8&8.3 &36.0& 18.8&34.9&13.9 \\
    \hline
    NRP & 51.3 & \textbf{48.4}&\textbf{47.3}&\textbf{50.3}&\textbf{50.8}&\textbf{50.2}&\textbf{51.4} \\
    \bottomrule[0.4mm]
\end{tabular}\vspace{-0.3em}
\label{tab:NRP-Classfication-Segmentation-Detection}
\end{table}

%% file: fig_tab_classification.tex
\begin{table*}[ht]
\centering\setlength{\tabcolsep}{7pt}
    \caption{\small Success rate (\emph{lower is better}) of BPDA \cite{carlini2017towards} and DIM$_{TI}$ \cite{dong2019evading} attacks against NRP. Res-v2-152 \cite{he2016identity} is combined with other purifier networks (ResG~\cite{ledig2017photo}, UNet \cite{ronneberger2015u}). Adversaries are then transferred to the naturally %(IncRes-v2) 
    and adversarially %(IncRes-v2\textsubscript{ens})
    trained models. NRP protects the backbone network even when the attacker tries to bypass using BPDA technique. (attack iterations: 10, $\epsilon \le 16$)} \vspace{-.6em}
 \scalebox{0.9}{
    \begin{tabular}{l|c|c|c|c|c|c|c|c}
    \toprule[0.4mm]
    \multirow{1}{*}{Source}&\multirow{1}{*}{Attack} & NRP  &Inc-v3&Inc-v4&IncRes-v2&Adv-v3&Inc-v3\textsubscript{ens3}&IncRes-v2\textsubscript{ens}\\
   \midrule[0.4mm]
    Res-v2-152 &DIM$_{TI}$ & \xmark &77.4&77.9&74.2&51.2&56.2&47.7\\
    \hline
    %     &   & \multicolumn{6}{c}{NRP is deployed}\\
    % \cline{3-8}
    ResG $\oplus$ Res-v2-152 & DIM$_{TI}$ $\oplus$ BPDA & \cmark &  29.7&26.2&19.6 & 22.3&22.1&16.1\\
    \hline
    UNet $\oplus$ Res-v2-152 & DIM$_{TI}$ $\oplus$ BPDA & \cmark &29.0&27.1& 19.5&26.9&27.7&18.8\\
    \bottomrule[0.4mm]
    \end{tabular}}\vspace{-0.3em}

    \label{tab:bpda_demo}
\end{table*}

\begin{figure*}[t]
\centering
  \begin{minipage}{.19\textwidth}
  	\centering
  \includegraphics[width=\linewidth,keepaspectratio]{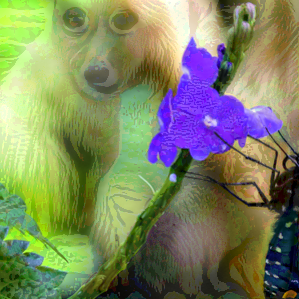}\
  \footnotesize Afghan Hound (0.73, \xmark)
  \end{minipage}
    \begin{minipage}{.19\textwidth}
  	\centering
  \includegraphics[width=\linewidth, keepaspectratio]{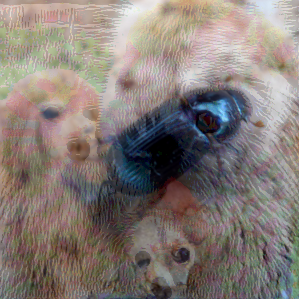}\
  \footnotesize Porcupine (0.64, \xmark)
  \end{minipage}
    \begin{minipage}{.19\textwidth}
  	\centering
  \includegraphics[width=\linewidth, keepaspectratio]{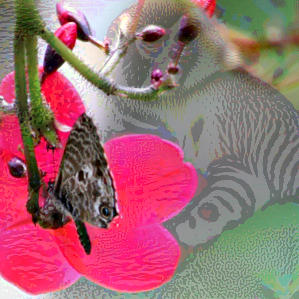}\
  \footnotesize Erythrocebus Patas (0.53, \xmark)
  \end{minipage}
    \begin{minipage}{.19\textwidth}
  	\centering
   \includegraphics[width=\linewidth, keepaspectratio]{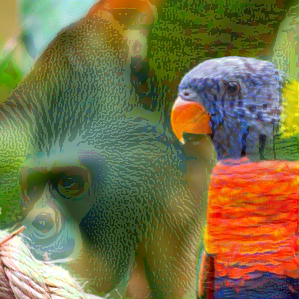}\
   \footnotesize  Guenon Monkey (0.77, \xmark)
  \end{minipage}
      \begin{minipage}{.19\textwidth}
  	\centering
  	\includegraphics[trim= 0 0 0 0mm, width=\linewidth, keepaspectratio]{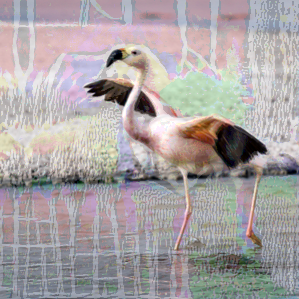}\
  	\footnotesize Crane (0.55, \xmark)
  \end{minipage}\vspace{0.05cm}
%Adversaries produced by NRD in feature space of a robust model \cite{engstrom2019adversarial}.  
  \\
  \begin{minipage}{.19\textwidth}
  	\centering
  	\footnotesize Monarch Butterfly (0.65, \cmark)
  \includegraphics[width=\linewidth,keepaspectratio]{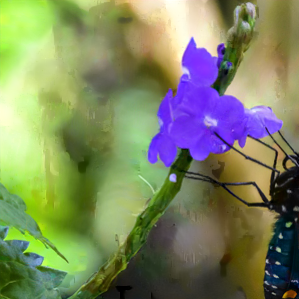}\
  \end{minipage}
    \begin{minipage}{.19\textwidth}
  	\centering
  	\footnotesize Dung Beetle (0.90, \cmark)
  \includegraphics[width=\linewidth, keepaspectratio]{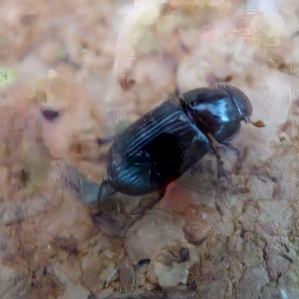}\
  \end{minipage}
    \begin{minipage}{.19\textwidth}
  	\centering
  	\footnotesize Lycaenid (0.94, \cmark)
  \includegraphics[width=\linewidth, keepaspectratio]{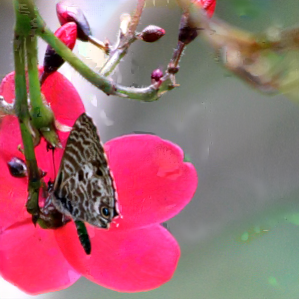}\
  \end{minipage}
    \begin{minipage}{.19\textwidth}
  	\centering
   \footnotesize Lorikeet (0.94, \cmark)
   \includegraphics[trim= 0 0 0 0mm, width=\linewidth, keepaspectratio]{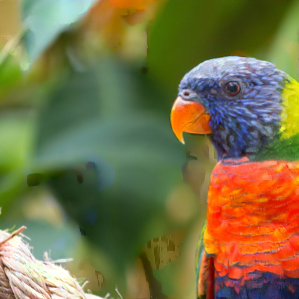}\
  \end{minipage}
      \begin{minipage}{.19\textwidth}
  	\centering
  	\footnotesize Flamingo (0.90, \cmark) 
  	\includegraphics[width=\linewidth, keepaspectratio]{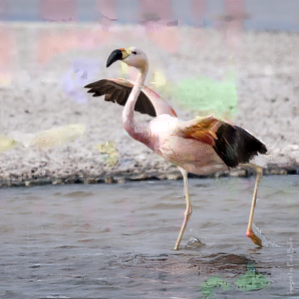}\
  \end{minipage}\vspace{0.05cm}
%Purified adversaries by NRP.
\vspace{-0.25cm}
  \caption{\small A visual illustration of NRP generalizability to different adversaries ($\epsilon \le 16$) (\emph{top:} attacked; \emph{bottom:} purified). Our method can clean  challenging adversarial patterns resulting from SSP applied to adversarially robust model \cite{engstrom2019adversarial}. %The noise is structured and  in-painted onto the original image. 
  Previous denoising methods are not designed for this type of structured noise. IncRes-v2\textsubscript{ens} backbone is used here. (see supplementary material for more examples)}
  \label{fig:NRP-removing-perturbation}
  \vspace{-0.8em}
\end{figure*}
\begin{figure*}[t]
\centering
  \begin{minipage}{0.24\textwidth}
  	\centering
    \includegraphics[trim= 0 0 0 0mm, width=\linewidth, keepaspectratio]{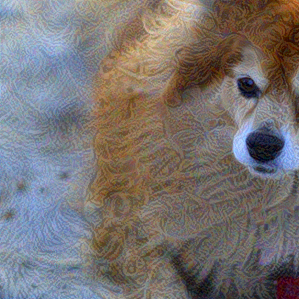}\
    \footnotesize DIM \cite{xie2018improving}: Welsh Springer (0.52, \xmark) 
  \end{minipage}
  \begin{minipage}{0.24\textwidth}
  	\centering
    \includegraphics[width=\linewidth, keepaspectratio]{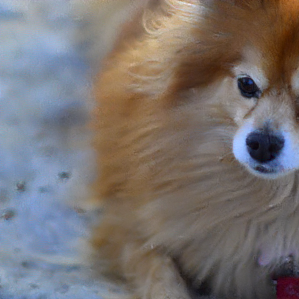}\
      \footnotesize Purified: Pomeranian (0.88, \cmark)
  \end{minipage}
  \begin{minipage}{0.24\textwidth}
  	\centering
    \includegraphics[trim= 0 0 0 0mm, width=\linewidth, keepaspectratio]{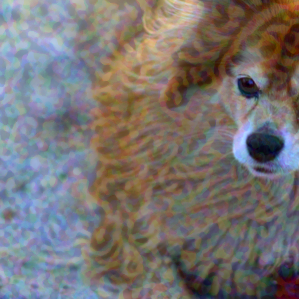}\
     \footnotesize DIM$_{TI}$ \cite{dong2019evading}: Cocker (0.71, \xmark)
  \end{minipage}
  \begin{minipage}{0.24\textwidth}
  	\centering
    \includegraphics[width=\linewidth, keepaspectratio]{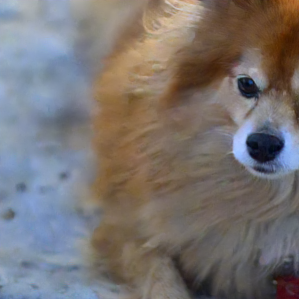}\
      \footnotesize Purified: Pomeranian (0.86, \cmark)
  \end{minipage}
  \vspace{-0.2cm}
  \caption{\small NRP successfully recovers diverse patterns from strongest black-box attacks ($l_\infty \le 16$). IncRes-v2\textsubscript{ens} used as backbone.}
\label{fig:NRP-classification-demo}
\end{figure*}
%##############################################
% Object detection figure
\begin{figure*}[t]
\centering
  \begin{minipage}{0.24\textwidth}
  	\centering
    \includegraphics[trim= 0 0 0 -2mm, width=\linewidth,keepaspectratio]{images/nrp_demo/000000022935.png}\
      \small CDA \cite{naseer2019cross}: Adversarial
  \end{minipage}
  \begin{minipage}{0.24\textwidth}
  	\centering
  	% left lower right up 
    \includegraphics[width=\linewidth,keepaspectratio]{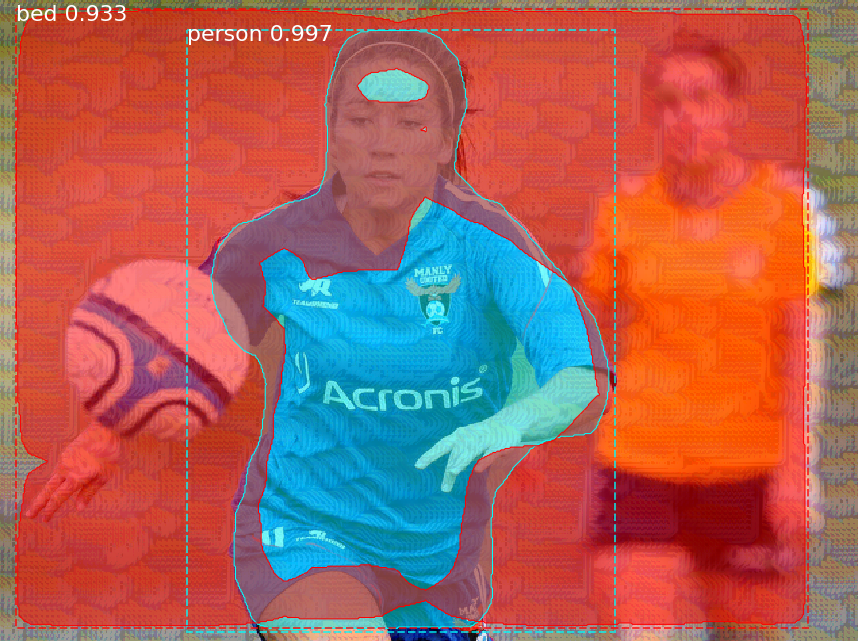}\
      \small Prediction for  Adversarial
  \end{minipage}
  \begin{minipage}{0.24\textwidth}
  	\centering
    \includegraphics[width=\linewidth, keepaspectratio]{images/nrp_demo/000000022935_cda_paintings_fixed.png}\
      \small Purified
  \end{minipage}
  \begin{minipage}{0.24\textwidth}
  	\centering
    \includegraphics[width=\linewidth, , keepaspectratio]{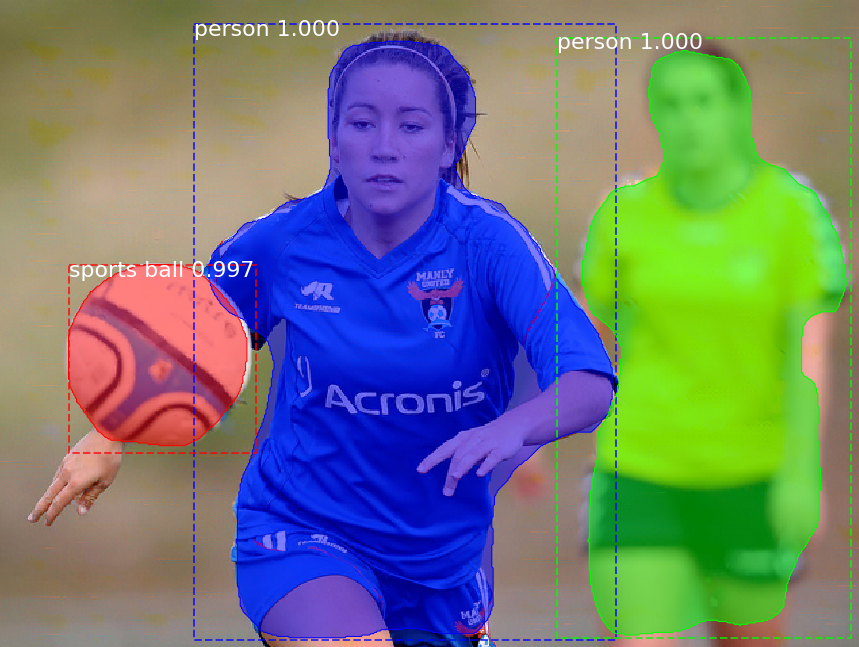}\
     \small Prediction for Purified
  \end{minipage} 
  \vspace{-0.2cm}
  \caption{\small NRP successfully removes perturbation generated by CDA\cite{naseer2019cross} ($\epsilon \le 16$) and stabilizes Mask-RCNN \cite{He2017MaskR} predictions.}
\label{fig:NRP-detection-demo}
\end{figure*}

%% file: ssp_attack_tables.tex
\begin{table}[!t]
\footnotesize\setlength{\tabcolsep}{2pt}
\caption{\small Cross-task SSP Attack: Pixel-level accuracy is shown for SegNet-Basic \cite{Badrinarayanan2017SegNetAD} on Camvid testset \cite{brostow2009semantic}, while mAP (with IoU = 0.5) is reported for Mask-RCNN.} \vspace{-.5em}
\centering
\scalebox{0.95}{
    \begin{tabular}{l|l|c|c|c}
        \toprule[.4mm]
        Problem & Method & No Attack & {SSP} ($l_\infty {\le} 8$) & SSP ($l_\infty {\le} 16$)\\
        %\cline{3-4} 
        %&  & $l_\infty \le 8$ & $l_\infty \le 16$   \\
        \midrule[0.4mm]
           
       Semantic Seg. &  SegNet \cite{Badrinarayanan2017SegNetAD} & 79.70 & 52.48 & 32.59  \\
        \midrule[0.4mm]  
      %  \multicolumn{4}{c}{\textbf{Instance Level Segmentation:} Mask-RCNN \cite{He2017MaskR}} \\ \midrule[0.4mm]  
      Instance Seg. &   Mask-RCNN \cite{He2017MaskR} & 56.8& 29.4 & 8.8  \\ \midrule[0.4mm]  
        % \multicolumn{4}{c}{\textbf{Object Detection:} Mask-RCNN \cite{He2017MaskR}} \\ \midrule[0.4mm]  
       Object Det. &  RetinaNet \cite{lin2018focal} &53.78  & 22.75 & 5.16  \\
       % \hline
       &  Mask-RCNN \cite{He2017MaskR} & 59.50 & 31.8 &  9.7 \\
        \bottomrule[0.4mm]
        \end{tabular}}\vspace{-1em}
\label{tab:SSP-Segmentation-Detection}
\end{table}

%##############################################
% Table: L_if transferability results (ImageNet-NIPS)
\begin{table*}[!htp]
\footnotesize
\caption{SSP as an attack for Classification. Top-1 (T-1) and Top-5 (T-5) accuracies are reported under untargeted $l_\infty$ adversarial attacks on ImageNet-NIPS with perturbation budget $l_\infty \le 16$. `$^*$' indicates white-box attacks.} \vspace{-2em}
\begin{center}
\begin{tabular}{l|p{14ex}<{\raggedright}|p{3.8ex}<{\centering} p{3.8ex}<{\centering}|p{3.8ex}<{\centering} p{3.8ex}<{\centering}|p{3.8ex}<{\centering} p{3.8ex}<{\centering}|p{3.8ex}<{\centering} p{3.8ex}<{\centering}|p{3.8ex}<{\centering} p{3.8ex}<{\centering}|p{3.8ex}<{\centering} p{3.8ex}<{\centering}|p{3.8ex}<{\centering} p{3.8ex}<{\centering}|p{3.8ex}<{\centering} c}
\toprule[0.4mm]
& &\multicolumn{10}{c|}{Naturally Trained}&\multicolumn{6}{c}{Adv. Trained}\\
\cline{3-18}
& Attack & \multicolumn{2}{c|}{Inc-v3}  & \multicolumn{2}{c|}{Inc-v4} & \multicolumn{2}{c|}{Res-152} & \multicolumn{2}{c|}{IncRes-v2} & \multicolumn{2}{c|}{VGG-19} & \multicolumn{2}{c|}{Adv-v3} & \multicolumn{2}{c|}{Inc-v3\textsubscript{ens3}} & \multicolumn{2}{c}{IncRes-v2\textsubscript{ens}} \\
\cline{3-18}
& & T-1 & T-5&T-1&T-5&T-1&T-5&T-1&T-5&T-1&T-5&T-1&T-5&T-1&T-5&T-1&T-5\\

\midrule[0.4mm]

\multirow{5}{*}{\rotatebox[origin=c]{90}{\parbox[c]{1.5cm}{\centering\texttt{Res-152}}}} & FGSM \cite{goodfellow2014explaining}&55.1&81.1&62.6&85.1&18.9$^*$&44.7$^*$&65.0&86.5&43.9&70.4&64.6&85.8&76.9&93.5&87.9&98.2 \\
& R-FGSM \cite{tramer2017ensemble}&60.8&84.3&68.4&88.1&14.6$^*$&40.3$^*$&71.9&90.3&55.8&71.4&74.8&92.3&81.1&96.0&87.1& 97.5\\
& I-FGSM \cite{Kurakin2016AdversarialEI}&80.9&96.7&85.3&97.8&0.9$^*$&10.8$^*$&93.1&98.8&75.9&94.8&89.2&99.2&90.5&97.9&94.6&99.5  \\
& MI-FGSM \cite{Dong_2018_CVPR}&38.9&72.7&44.8&76.5&0.6$^*$&2.9$^*$&47.7&79.6&42.1&71.8&67.0&89.9&69.4&93.3&81.5&96.4 \\
& TAP \cite{Zhou_2018_ECCV} &48.2&-&55.7&-&7.6$^*$&-&55.2&-&-&-&49.2&-&57.8&-&64.1&-\\
&DIM \cite{xie2018improving}&15.9&44.0&17.3&48.4&0.8$^*$&3.0$^*$&20.0&50.2&25.6&56.3&55.8&82.8&54.9&84.2&71.5&93.1  \\
\hline

\multirow{7}{*}{\rotatebox[origin=c]{90}{\parbox[c]{1.5cm}{\centering\texttt{VGG16}}}} &FGSM \cite{goodfellow2014explaining}&32.6&58.6&38.4&62.6&38.5&66.3&44.5&68.5&8.8&25.1&51.7&75.3&54.9&81.7&70.8&90.7\\
&R-FGSM \cite{tramer2017ensemble}&44.4&69.5&47.6&75.1&51.1&78.8&56.4&78.8&11.2&31.8&65.5&87.4&66.7&89.2&77.5&93.6\\
&I-FGSM \cite{Kurakin2016AdversarialEI}&69.2&93.0&75.2&93.7&79.0&96.2&85.6&96.8&14.4&49.3&83.5&97.7&83.9&96.7&92.1&98.8\\
&MI-FGSM \cite{Dong_2018_CVPR}&20.4&45.0&19.7&43.2&25.2&53.8&26.8&53.8&1.5&12.1&43.0&70.9&42.0&72.7&62.0&86.8\\
& TAP \cite{Zhou_2018_ECCV} & 23.9 & - & 28.1&-&23.9&-&32.3&-&-&-&38.8&-&41.9&-&63.8&-\\
&DIM \cite{xie2018improving}& 14.7&38.8&16.6&39.0&21.0&48.0&21.5&45.7&\textbf{0.6}&7.6&35.8&65.8&\textbf{31.8}&60.8&\textbf{53.7}&79.5 \\
% \cline{2-18}
&FFF \cite{mopuri-bmvc-2017} &61.7&80.7&60.8&78.7&72.8&90.1&76.1&90.1&44.0&68.0&79.6&93.1&83.1&93.1&92.8&98.5  \\
\cline{2-18}
&SSP&\textbf{5.3}&\textbf{11.0}&\textbf{5.9}&\textbf{11.9}&\textbf{16.5}&\textbf{29.5}&\textbf{14.1}&\textbf{25.5}&2.7&\textbf{6.8}&\textbf{25.9}&\textbf{43.2}&40.2&\textbf{58.3}&58.0&\textbf{75.0}\\
\bottomrule[0.4mm]
\end{tabular}
\end{center}
\vspace{-3ex}
\label{tab:att-nips}
\end{table*}

%% file: NRP_Paper.bbl
\begin{thebibliography}{10}\itemsep=-1pt

\bibitem{arjovsky2017wasserstein}
Martin Arjovsky, Soumith Chintala, and L{\'e}on Bottou.
\newblock Wasserstein gan.
\newblock {\em arXiv preprint arXiv:1701.07875}, 2017.

\bibitem{athalye2018obfuscated}
Anish Athalye, Nicholas Carlini, and David~A. Wagner.
\newblock Obfuscated gradients give a false sense of security: Circumventing
  defenses to adversarial examples.
\newblock In {\em International Conference on Machine Learning (ICML)}, 2018.

\bibitem{athalye2017synthesizing}
Anish Athalye, Logan Engstrom, Andrew Ilyas, and Kevin Kwok.
\newblock Synthesizing robust adversarial examples.
\newblock In {\em International Conference on Machine Learning (ICML)}, 2017.

\bibitem{Badrinarayanan2017SegNetAD}
Vijay Badrinarayanan, Alex Kendall, and Roberto Cipolla.
\newblock Segnet: A deep convolutional encoder-decoder architecture for image
  segmentation.
\newblock {\em IEEE Transactions on Pattern Analysis and Machine Intelligence},
  39:2481--2495, 2017.

\bibitem{brostow2009semantic}
Gabriel~J Brostow, Julien Fauqueur, and Roberto Cipolla.
\newblock Semantic object classes in video: A high-definition ground truth
  database.
\newblock {\em Pattern Recognition Letters}, 30(2):88--97, 2009.

\bibitem{carlini2017towards}
Nicholas Carlini and David Wagner.
\newblock Towards evaluating the robustness of neural networks.
\newblock In {\em 2017 IEEE Symposium on Security and Privacy (SP)}, pages
  39--57. IEEE, 2017.

\bibitem{Nips-data}
NeurIPS Challenge.
\newblock
  \url{https://www.kaggle.com/c/nips-2017-defense-against-adversarial-attack/data}.
\newblock {\em Kaggle}, 2017.

\bibitem{cohen2019certified}
Jeremy~M Cohen, Elan Rosenfeld, and J~Zico Kolter.
\newblock Certified adversarial robustness via randomized smoothing.
\newblock {\em arXiv preprint arXiv:1902.02918}, 2019.

\bibitem{Dong_2018_CVPR}
Yinpeng Dong, Fangzhou Liao, Tianyu Pang, Hang Su, Jun Zhu, Xiaolin Hu, and
  Jianguo Li.
\newblock Boosting adversarial attacks with momentum.
\newblock In {\em The IEEE Conference on Computer Vision and Pattern
  Recognition (CVPR)}, June 2018.

\bibitem{dong2019evading}
Yinpeng Dong, Tianyu Pang, Hang Su, and Jun Zhu.
\newblock Evading defenses to transferable adversarial examples by
  translation-invariant attacks.
\newblock In {\em Proceedings of the IEEE Computer Society Conference on
  Computer Vision and Pattern Recognition}, 2019.

\bibitem{engstrom2019a}
Logan Engstrom, Justin Gilmer, Gabriel Goh, Dan Hendrycks, Andrew Ilyas,
  Aleksander Madry, Reiichiro Nakano, Preetum Nakkiran, Shibani Santurkar,
  Brandon Tran, Dimitris Tsipras, and Eric Wallace.
\newblock A discussion of 'adversarial examples are not bugs, they are
  features'.
\newblock {\em Distill}, 2019.

\bibitem{engstrom2019adversarial}
Logan Engstrom, Andrew Ilyas, Shibani Santurkar, Dimitris Tsipras, Brandon
  Tran, and Aleksander Madry.
\newblock Adversarial robustness as a prior for learned representations, 2019.

\bibitem{geirhos2018imagenettrained}
Robert Geirhos, Patricia Rubisch, Claudio Michaelis, Matthias Bethge, Felix~A.
  Wichmann, and Wieland Brendel.
\newblock Imagenet-trained {CNN}s are biased towards texture; increasing shape
  bias improves accuracy and robustness.
\newblock In {\em International Conference on Learning Representations}, 2019.

\bibitem{goodfellow2014explaining}
Ian Goodfellow, Jonathon Shlens, and Christian Szegedy.
\newblock Explaining and harnessing adversarial examples.
\newblock In {\em International Conference on Learning Representations (ICRL)},
  2015.

\bibitem{Kurakin2016AdversarialEI}
Ian~J Goodfellow, Jonathon Shlens, and Christian Szegedy.
\newblock Adversarial examples in the physical world.
\newblock In {\em International Conference on Learning Representations (ICRL)},
  2017.

\bibitem{guo2017countering}
Chuan Guo, Mayank Rana, Moustapha Ciss{\'e}, and Laurens van~der Maaten.
\newblock Countering adversarial images using input transformations.
\newblock In {\em International Conference on Learning Representations (ICRL)},
  2017.

\bibitem{guo2018countering}
Chuan Guo, Mayank Rana, Moustapha Cisse, and Laurens van~der Maaten.
\newblock Countering adversarial images using input transformations.
\newblock In {\em International Conference on Learning Representations}, 2018.

\bibitem{He2017MaskR}
Kaiming He, Georgia Gkioxari, Piotr Doll{\'a}r, and Ross~B. Girshick.
\newblock Mask r-cnn.
\newblock {\em 2017 IEEE International Conference on Computer Vision (ICCV)},
  pages 2980--2988, 2017.

\bibitem{he2016deep}
Kaiming He, Xiangyu Zhang, Shaoqing Ren, and Jian Sun.
\newblock Deep residual learning for image recognition.
\newblock In {\em Proceedings of the IEEE conference on computer vision and
  pattern recognition}, pages 770--778, 2016.

\bibitem{he2016identity}
Kaiming He, Xiangyu Zhang, Shaoqing Ren, and Jian Sun.
\newblock Identity mappings in deep residual networks.
\newblock In {\em European conference on computer vision}, pages 630--645.
  Springer, 2016.

\bibitem{Iandola2017SqueezeNetAA}
Forrest~N. Iandola, Matthew~W. Moskewicz, Khalid Ashraf, Song Han, William~J.
  Dally, and Kurt Keutzer.
\newblock Squeezenet: Alexnet-level accuracy with 50x fewer parameters and <1mb
  model size.
\newblock {\em ArXiv}, abs/1602.07360, 2017.

\bibitem{ilyas2019adversarial}
Andrew Ilyas, Shibani Santurkar, Dimitris Tsipras, Logan Engstrom, Brandon
  Tran, and Aleksander Madry.
\newblock Adversarial examples are not bugs, they are features.
\newblock {\em arXiv preprint arXiv:1905.02175}, 2019.

\bibitem{jolicoeur2018relativistic}
Alexia Jolicoeur-Martineau.
\newblock The relativistic discriminator: a key element missing from standard
  gan.
\newblock {\em arXiv preprint arXiv:1807.00734}, 2018.

\bibitem{Krizhevsky2012ImageNetCW}
Alex Krizhevsky, Ilya Sutskever, and Geoffrey~E. Hinton.
\newblock Imagenet classification with deep convolutional neural networks.
\newblock {\em Commun. ACM}, 60:84--90, 2012.

\bibitem{Kupyn_2018_CVPR}
Orest Kupyn, Volodymyr Budzan, Mykola Mykhailych, Dmytro Mishkin, and Jiří
  Matas.
\newblock Deblurgan: Blind motion deblurring using conditional adversarial
  networks.
\newblock In {\em The IEEE Conference on Computer Vision and Pattern
  Recognition (CVPR)}, June 2018.

\bibitem{kurakin2016adversarial}
Alexey Kurakin, Ian Goodfellow, and Samy Bengio.
\newblock Adversarial machine learning at scale.
\newblock {\em arXiv preprint arXiv:1611.01236}, 2016.

\bibitem{ledig2017photo}
Christian Ledig, Lucas Theis, Ferenc Husz{\'a}r, Jose Caballero, Andrew
  Cunningham, Alejandro Acosta, Andrew Aitken, Alykhan Tejani, Johannes Totz,
  Zehan Wang, et~al.
\newblock Photo-realistic single image super-resolution using a generative
  adversarial network.
\newblock In {\em Proceedings of the IEEE conference on computer vision and
  pattern recognition}, pages 4681--4690, 2017.

\bibitem{Liao_2018_CVPR}
Fangzhou Liao, Ming Liang, Yinpeng Dong, Tianyu Pang, Xiaolin Hu, and Jun Zhu.
\newblock Defense against adversarial attacks using high-level representation
  guided denoiser.
\newblock In {\em The IEEE Conference on Computer Vision and Pattern
  Recognition (CVPR)}, June 2018.

\bibitem{Liao2017DefenseAA}
Fangzhou Liao, Ming Liang, Yinpeng Dong, Tianyu Pang, Jun Zhu, and Xiaolin Hu.
\newblock Defense against adversarial attacks using high-level representation
  guided denoiser.
\newblock {\em 2018 IEEE/CVF Conference on Computer Vision and Pattern
  Recognition}, pages 1778--1787, 2017.

\bibitem{lin2018focal}
Tsung-Yi Lin, Priyal Goyal, Ross Girshick, Kaiming He, and Piotr Doll{\'a}r.
\newblock Focal loss for dense object detection.
\newblock {\em IEEE transactions on pattern analysis and machine intelligence},
  2018.

\bibitem{lin2014microsoft}
Tsung-Yi Lin, Michael Maire, Serge Belongie, James Hays, Pietro Perona, Deva
  Ramanan, Piotr Doll{\'a}r, and C~Lawrence Zitnick.
\newblock Microsoft coco: Common objects in context.
\newblock In {\em European conference on computer vision}, pages 740--755.
  Springer, 2014.

\bibitem{maaten2008visualizing}
Laurens van~der Maaten and Geoffrey Hinton.
\newblock Visualizing data using t-sne.
\newblock {\em Journal of machine learning research}, 9(Nov):2579--2605, 2008.

\bibitem{madry2018towards}
Aleksander Madry, Aleksandar Makelov, Ludwig Schmidt, Dimitris Tsipras, and
  Adrian Vladu.
\newblock Towards deep learning models resistant to adversarial attacks.
\newblock In {\em International Conference on Learning Representations}, 2018.

\bibitem{mopuri-bmvc-2017}
Konda~Reddy Mopuri, Utsav Garg, and R~Venkatesh Babu.
\newblock Fast feature fool: A data independent approach to universal
  adversarial perturbations.
\newblock In {\em Proceedings of the British Machine Vision Conference
  ({BMVC})}, 2017.

\bibitem{mustafa2019image}
Aamir Mustafa, Salman~H Khan, Munawar Hayat, Jianbing Shen, and Ling Shao.
\newblock Image super-resolution as a defense against adversarial attacks.
\newblock {\em arXiv preprint arXiv:1901.01677}, 2019.

\bibitem{nakano2019robuststyle}
Reiichiro Nakano.
\newblock Neural style transfer with adversarially robust classifiers, Jun
  2019.

\bibitem{naseer2019cross}
Muzammal Naseer, Salman~H Khan, Harris Khan, Fahad~Shahbaz Khan, and Fatih
  Porikli.
\newblock Cross-domain transferability of adversarial perturbations.
\newblock {\em Advances in Neural Information Processing Systems}, 2019.

\bibitem{ronneberger2015u}
Olaf Ronneberger, Philipp Fischer, and Thomas Brox.
\newblock U-net: Convolutional networks for biomedical image segmentation.
\newblock In {\em International Conference on Medical image computing and
  computer-assisted intervention}, pages 234--241. Springer, 2015.

\bibitem{ILSVRC15}
Olga Russakovsky, Jia Deng, Hao Su, Jonathan Krause, Sanjeev Satheesh, Sean Ma,
  Zhiheng Huang, Andrej Karpathy, Aditya Khosla, Michael Bernstein,
  Alexander~C. Berg, and Li Fei-Fei.
\newblock {ImageNet Large Scale Visual Recognition Challenge}.
\newblock {\em International Journal of Computer Vision (IJCV)},
  115(3):211--252, 2015.

\bibitem{shafahi2019adversarial}
Ali Shafahi, Mahyar Najibi, Amin Ghiasi, Zheng Xu, John Dickerson, Christoph
  Studer, Larry~S Davis, Gavin Taylor, and Tom Goldstein.
\newblock Adversarial training for free!
\newblock {\em arXiv preprint arXiv:1904.12843}, 2019.

\bibitem{shen2017ape}
Shiwei Shen, Guoqing Jin, Ke Gao, and Yongdong Zhang.
\newblock Ape-gan: Adversarial perturbation elimination with gan.
\newblock {\em arXiv preprint arXiv:1707.05474}, 2017.

\bibitem{Simonyan14verydeep}
Karen Simonyan and Andrew Zisserman.
\newblock Very deep convolutional networks for large-scale image recognition,
  2014.

\bibitem{Su2018}
Dong Su, Huan Zhang, Hongge Chen, Jinfeng Yi, Pin-Yu Chen, and Yupeng Gao.
\newblock Is robustness the cost of accuracy? -- a comprehensive study on the
  robustness of 18 deep image classification models.
\newblock In Vittorio Ferrari, Martial Hebert, Cristian Sminchisescu, and Yair
  Weiss, editors, {\em Computer Vision -- ECCV 2018}, pages 644--661, Cham,
  2018. Springer International Publishing.

\bibitem{szegedy2017inception}
Christian Szegedy, Sergey Ioffe, Vincent Vanhoucke, and Alexander~A Alemi.
\newblock Inception-v4, inception-resnet and the impact of residual connections
  on learning.
\newblock In {\em AAAI}, volume~4, page~12, 2017.

\bibitem{szegedy2016rethinking}
Christian Szegedy, Vincent Vanhoucke, Sergey Ioffe, Jon Shlens, and Zbigniew
  Wojna.
\newblock Rethinking the inception architecture for computer vision.
\newblock In {\em Proceedings of the IEEE Conference on Computer Vision and
  Pattern Recognition}, pages 2818--2826, 2016.

\bibitem{szegedy2013intriguing}
Christian Szegedy, Wojciech Zaremba, Ilya Sutskever, Joan Bruna, Dumitru Erhan,
  Ian Goodfellow, and Rob Fergus.
\newblock Intriguing properties of neural networks.
\newblock In {\em International Conference on Learning Representations (ICRL)},
  2014.

\bibitem{NIPS-r3}
Anil Thomas and Oguz Elibol.
\newblock Defense against adversarial attacks-3rd place.
\newblock
  \url{https://github.com/anlthms/nips-2017/blob/master/poster/defense.pdf},
  2017.

\bibitem{tramer2017ensemble}
Florian Tram{\`e}r, Alexey Kurakin, Nicolas Papernot, Dan Boneh, and Patrick
  McDaniel.
\newblock Ensemble adversarial training: Attacks and defenses.
\newblock In {\em International Conference on Learning Representations (ICRL)},
  2018.

\bibitem{wang2019edvr}
Xintao Wang, Kelvin~C.K. Chan, Ke Yu, Chao Dong, and Chen~Change Loy.
\newblock Edvr: Video restoration with enhanced deformable convolutional
  networks.
\newblock In {\em The IEEE Conference on Computer Vision and Pattern
  Recognition (CVPR) Workshops}, June 2019.

\bibitem{xie2017mitigating}
Cihang Xie, Jianyu Wang, Zhishuai Zhang, Zhou Ren, and Alan Yuille.
\newblock Mitigating adversarial effects through randomization.
\newblock {\em arXiv preprint arXiv:1711.01991}, 2017.

\bibitem{xie2018mitigating}
Cihang Xie, Jianyu Wang, Zhishuai Zhang, Zhou Ren, and Alan Yuille.
\newblock Mitigating adversarial effects through randomization.
\newblock In {\em International Conference on Learning Representations}, 2018.

\bibitem{xie2019feature}
Cihang Xie, Yuxin Wu, Laurens van~der Maaten, Alan~L Yuille, and Kaiming He.
\newblock Feature denoising for improving adversarial robustness.
\newblock In {\em Proceedings of the IEEE Conference on Computer Vision and
  Pattern Recognition}, pages 501--509, 2019.

\bibitem{xie2018improving}
Cihang Xie, Zhishuai Zhang, Jianyu Wang, Yuyin Zhou, Zhou Ren, and Alan Yuille.
\newblock Improving transferability of adversarial examples with input
  diversity.
\newblock {\em arXiv preprint arXiv:1803.06978}, 2018.

\bibitem{xu2015empirical}
Bing Xu, Naiyan Wang, Tianqi Chen, and Mu Li.
\newblock Empirical evaluation of rectified activations in convolutional
  network.
\newblock {\em arXiv preprint arXiv:1505.00853}, 2015.

\bibitem{Zantedeschi2017EfficientDA}
Valentina Zantedeschi, Maria-Irina Nicolae, and Ambrish Rawat.
\newblock Efficient defenses against adversarial attacks.
\newblock {\em ArXiv}, abs/1707.06728, 2017.

\bibitem{zhang2019defense}
Haichao Zhang and Jianyu Wang.
\newblock Defense against adversarial attacks using feature scattering-based
  adversarial training.
\newblock {\em arXiv preprint arXiv:1907.10764}, 2019.

\bibitem{Zhang2018TheUE}
Richard Zhang, Phillip Isola, Alexei~A. Efros, Eli Shechtman, and Oliver Wang.
\newblock The unreasonable effectiveness of deep features as a perceptual
  metric.
\newblock {\em 2018 IEEE/CVF Conference on Computer Vision and Pattern
  Recognition}, pages 586--595, 2018.

\bibitem{Zhou_2018_ECCV}
Wen Zhou, Xin Hou, Yongjun Chen, Mengyun Tang, Xiangqi Huang, Xiang Gan, and
  Yong Yang.
\newblock Transferable adversarial perturbations.
\newblock In {\em The European Conference on Computer Vision (ECCV)}, September
  2018.

\end{thebibliography}
